\documentclass{article}
\pdfoutput=1

\usepackage[nonatbib,final]{neurips_2020}

\usepackage[utf8]{inputenc} %
\usepackage[T1]{fontenc}    %
\usepackage{hyperref}       %
\usepackage{url}            %
\usepackage{booktabs}       %
\usepackage{amsfonts}       %
\usepackage{nicefrac}       %
\usepackage{microtype}      %

\usepackage{macros}

\title{BoxE: A Box Embedding Model for\\ Knowledge Base Completion}

\author{%
  Ralph Abboud, {\.I}smail {\.I}lkan Ceylan, Thomas Lukasiewicz, Tommaso Salvatori \\
  Department of Computer Science\\
  University of Oxford, UK\\
  \texttt{\{firstame.lastname\}@cs.ox.ac.uk} \\
}

\begin{document}

\maketitle

\begin{abstract}
Knowledge base completion (KBC) aims to automatically infer missing facts by exploiting information already present in a knowledge base (KB). A promising approach for KBC is to embed knowledge into latent spaces and make predictions from learned embeddings.
However, existing embedding models are subject to at least one of the following limitations: (1)~theoretical \emph{inexpressivity}, (2)~lack of support for prominent \emph{inference patterns} (e.g., hierarchies), (3)~lack of support for KBC over \emph{higher-arity} relations, and (4)~lack of support for incorporating \emph{logical rules}. 
Here, we propose  a \emph{spatio-translational} embedding model, 
called \emph{BoxE}, that simultaneously addresses  all these limitations. BoxE embeds entities as \emph{points}, and relations as a set of \emph{hyper-rectangles} (or \emph{boxes}), which spatially characterize basic logical properties. 
This seemingly simple abstraction yields a fully expressive model offering a natural encoding for many desired logical properties. BoxE can both \emph{capture} and \emph{inject} rules from rich classes  of rule languages, going well beyond individual inference patterns. 
By design, BoxE naturally applies to higher-arity KBs. 
We conduct a detailed experimental analysis, and show that BoxE achieves state-of-the-art performance, both on benchmark knowledge graphs and on more general KBs, and we empirically show the power of integrating logical rules. 
\end{abstract}
\section{Introduction}

Knowledge bases (KBs) are fundamental means for {representing}, {storing}, and {processing} information, and are widely used to enhance the \emph{reasoning} and \emph{learning} capabilities of modern information systems.
KBs can be viewed as a collection of facts of the form $r(e_1, \ldots,e_n)$, which represent a relation $r$ between the entities $e_1, \ldots, e_n$, and {knowledge graphs (KGs)} as a special case, where all the relations are binary (i.e., composed of two entities).
KBs such as YAGO \cite{MahdisoltaniBS15}, NELL \cite{MitchellBCM18}, Knowledge Vault \cite{GoogleVault}, and Freebase \cite{BollackerCT07} contain millions of facts, and are increasingly important in academia and industry, for applications such as question answering \cite{BordesCW14}, recommender systems \cite{WangZWZLXG18}, information retrieval \cite{XiongPC17}, and natural language processing \cite{YangM17}.

KBs are, however, highly \emph{incomplete}, which makes their downstream use %
more challenging. For instance, 71\% of individuals in Freebase lack a connection to a place of birth \cite{WestGMSGL14}.
\emph{Knowledge base completion (KBC)}, aiming at automatically inferring missing facts in a KB by exploiting the already present information, has thus become a focal point of research.
One prominent approach for KBC is to learn \emph{embeddings} for entities and relations in a latent space such that these embeddings, once learned from known facts, can be used to \emph{score} the plausibility of unknown facts. 

Currently, the main embedding approaches for KBC are translational models \cite{TransE-NIPS13, RotatE-ICLR19}, which score facts based on distances in the embedding space, bilinear models \cite{ComplEx-ICML16, DistMult-ICLR15, TuckER},  which learn embeddings that factorize the truth tensor of a knowledge base, and neural models \cite{ConvE-AAAI18, SocherCMN13, KBGAT-ACL19},  which score facts using dedicated neural architectures. 
Each of these models suffer from limitations, most of which are well-known. 
Translational models, for instance, are theoretically inexpressive, i.e., cannot provably fit an arbitrary KG. 
Furthermore, none of these models can capture simple sets of \emph{logical rules}: even capturing a simple relational hierarchy goes beyond the current capabilities of most existing models \cite{Gutirrez18}.
This also makes it difficult to inject background knowledge (i.e., schematic knowledge), in the form of logical rules, into the model to  improve KBC performance.
Additionally, existing KBC models are primarily designed for KGs, and thus do not naturally extend to KBs with \emph{higher-arity} relations, involving 3 or more entities, e.g.,  $\mathsf{DegreeFrom(Turing,PhD,Princeton)}$ \cite{Fatemi19}, which hinders their applicability. Higher-arity relations are prevalent in modern KBs such as Freebase \cite{Wen16}, and cannot always be reduced to a KG without loss of information \cite{Fatemi19}. 
Despite the rich landscape for KBC, no existing model currently offers a solution to all these limitations.

In this paper, we address these problems by encoding relations as explicit regions in the embedding space, where logical properties such as relation subsumption and disjointness can naturally be analyzed and inferred. Specifically, we present \emph{BoxE}, a \emph{spatio-translational} box embedding model, which models relations as sets of $d-$dimensional boxes (corresponding to classes), and entities as $d-$dimensional points. 
Facts are scored based on the positions of entity embeddings with respect to relation boxes. Our contributions can be summarized as follows:
\begin{itemize}[--,leftmargin=8pt]
	\item We introduce BoxE and show that this model achieves state-of-the-art performance on both \emph{knowledge graph completion} and \emph{knowledge base completion} tasks across multiple datasets.
	
	\item We show that BoxE is fully expressive, a first for translation-based models, to our knowledge. %
		
	\item We comprehensively analyze the inductive capacity of BoxE in terms of generalized inference patterns and rule languages, and show that BoxE can capture a rich rule language.

	\item We prove that BoxE additionally supports \emph{injecting} a rich language of logical rules, and empirically show on a subset of NELL \cite{MitchellBCM18}, that this can significantly improve KBC performance.
\end{itemize}

All proofs for theorems, as well as additional experiments and experimental details, can be found in the appendix of this paper.

\section{Knowledge Base Completion: Problem, Properties, and Evaluation}
In this section, we define knowledge bases and the problem of knowledge base completion (KBC). We also give an overview of standard approaches for evaluating KBC models.

Consider a \emph{relational vocabulary}, which consists of a finite set $\Ebf$ of \emph{entities} and a finite set $\Rbf$ of \emph{relations}. A \emph{fact} (also called \emph{atom}) is of the form $r(e_1, \ldots, e_n)$, where $r \in \Rbf$ is an $n$-ary relation, and $e_i \in \Ebf$ are entities. 
A \emph{knowledge base (KB)} is a finite set of facts, and a \emph{knowledge graph (KG)} is a KB with only binary relations. In KGs, facts are also known as \emph{triples}, and are of the form $r(e_h,e_t)$,  with a \emph{head} entity~$e_h$ and a \emph{tail} entity $e_t$. \emph{Knowledge base completion~(KBC)} (resp., knowledge graph completion (KGC)) is the task of accurately predicting new facts from existing facts in a KB (resp., KG). KBC models are analyzed by means of (i)~an \emph{experimental evaluation} on existing benchmarks, (ii)~their model \emph{expressiveness}, and (iii) the set of \emph{inference patterns} that they can capture.

\textbf{Experimental evaluation.} To evaluate KBC models empirically, \emph{true facts} from the test set of a KB and \emph{corrupted facts}, generated from the test set, are used.
A corrupted fact is obtained by replacing one of the entities in a fact from the KB with a new entity: given a fact $r(e_1, \ldots, e_i, \ldots, e_n)$ from the KB, a corrupted fact is a fact $r(e_1,\ldots,e_i',\ldots,e_n)$ that does \emph{not} occur in the training, validation, or test set.
KBC models define a \emph{scoring function} over facts, and are optimized to score true facts higher than corrupted facts. KBC performance is evaluated using metrics \cite{TransE-NIPS13} such as mean rank (MR), the average rank of facts against their corrupted counterparts, mean reciprocal rank (MRR), their average inverse rank (i.e., 1/rank), and  Hits@K, the proportion of facts with rank at most K. %

\textbf{Expressiveness.} A KBC model $\Mmc$ is  \emph{fully expressive} if, for any given disjoint sets of \emph{true} and \emph{false} facts, there exists a parameter configuration for $\Mmc$ such that $\Mmc$ accurately classifies all the given facts. 
Intuitively, a fully expressive model can capture any knowledge base configuration, but this does not necessarily correlate with inductive capacity: fully expressive models can merely memorize training data and generalize poorly. Conversely, a model that is not fully expressive can fail to fit its training set properly, and thus can underfit. Hence, it is important to develop models that are jointly fully expressive and capture prominent and common inference patterns.

\textbf{Inference patterns.}
Inference patterns are a common means to formally analyze the generalization ability of KBC systems. Briefly, an \emph{inference pattern} is a specification of a logical property that may exist in a KB, which, if learned, enables further principled inferences from existing KB facts.
One well-known example inference pattern is \emph{symmetry}, which specifies that when a fact $r(e_1, e_2)$ holds, then $r(e_2, e_1)$ also holds. If a model learns a symmetry pattern for $r$, then it can automatically predict facts in the symmetric closure of $r$, thus providing a strong inductive bias. 
We present some prominent inference patterns in detail in \Cref{sec:BoxEProps}, and also in \Cref{tab:infPat}. Intuitively, inference patterns captured by a model serve as an indication of its \emph{inductive capacity}.

\section{Related Work}

In this section, we give an overview of closely related embedding methods for KBC/KGC  and existing region-based embedding models. We exclude neural models \cite{ConvE-AAAI18,EMLP-NIPS13,KBGAT-ACL19}, as these models are challenging to analyze, both from an expressiveness and inductive capacity perspective. 

\textbf{Translational models.} Translational models represent entities as points in a high-dimensional vector space and relations as translations in this space. The seminal translational model is TransE \cite{TransE-NIPS13}, where a relation $r$, modeled by a vector $\bm{r}$, holds between $e_1$ and $e_2$ iff $\bm{e_1} + \bm{r} = \bm{e_2}$. However, TransE is not fully expressive, cannot capture \emph{one-to-many}, \emph{many-to-one},  \emph{many-to-many}, and symmetric relations, and can only handle binary facts. 
This motivated extensions \cite{TransH-AAAI14, TransR-AAAI15,TranSparse-AAAI16,TransF-KR16}, which each address some, but not all,  these limitations. 
Beyond translations, RotatE \cite{RotatE-ICLR19} uses rotations to model relations, and thus can model symmetric relations with rotations of angle $\theta = \pm \pi$, but is otherwise as limited as TransE. 
Translational models are interpretable and can capture various inference patterns, but no known translational model is fully expressive.

\textbf{Bilinear models.} Bilinear models capture relations as a bilinear product between entity and relation embeddings. RESCAL \cite{RESCAL-ICML11} represents a relation $r$ as a full-rank $d \times d$ matrix $M$, and entities as $d$-dimensional vectors $\bm{e}$. DistMult \cite{DistMult-ICLR15} simplifies RESCAL by making $M$ diagonal, but cannot capture non-symmetric relations.
ComplEx \cite{ComplEx-ICML16} defines a diagonal $M$ with complex numbers to capture anti-symmetry. SimplE \cite{SimplE-NeurIPS18} and TuckER \cite{TuckER} build on canonical polyadic (CP) \cite{hitchcock1927expression} and Tucker decomposition \cite{tucker1966some}, respectively. TuckER subsumes RESCAL, its adaptations, and SimplE~\cite{TuckER}. 
Generally, all bilinear models except DistMult are fully expressive, but they are less interpretable compared to translational models.

\textbf{Higher-arity KBC.}
KBs can encode knowledge that cannot be encoded in a KG \cite{Fatemi19}. Hence, models such as HSimplE \cite{Fatemi19}, m-TransH \cite{Wen16}, m-DistMult, and m-CP \cite{Fatemi19} are proposed as generalizations of SimplE, TransH \cite{TransH-AAAI14}, DistMult, and CP, respectively.  HypE \cite{Fatemi19} tackles higher-arity KBC through convolutions. 
Generalizations to TuckER, namely,  m-TuckER and GETD \cite{Liu20}, are also proposed, but these do not apply to KBs with different-arity relations.
For most existing KGC models, there are conceptual and practical challenges (e.g., scalability) against generalizing them to KBC.

\textbf{Region-based models.}
Region-based models explicitly define regions %
in the embedding space where an output property (e.g., membership to a class) holds. 
For instance, bounded axis-aligned hyper-rectangles (boxes) \cite{Box-ACL18,SubramanianC18,LiVZBM19} are used for entity classification to define class regions and hierarchies, in which entity point embeddings  appear. 
As boxes naturally represent sets of objects, they are also used to represent answer sets in the Query2Box query answering system~\cite{Ren20}. Query2Box can be applied to KBC but reduces to a translational model with a box correctness region for tail entities.
Furthermore, entity classification approaches cannot be scalably generalized to KBC, as this would involve introducing an embedding per entity tuple.

\section{Box Embeddings for Knowledge Base Completion}
\label{sec:BoxE}
In this section, we introduce an embedding model for KBC, 
called \emph{BoxE}, that encodes relations as axis-aligned \emph{hyper-rectangles} (or boxes) and entities as \emph{points} in the $d$-dimensional Euclidian space.

\textbf{Representation.} In BoxE, every entity $e_i \in \Ebf$ is represented by two vectors $\bm{e_i}, \bm{b_i} \in \mathbb{R}^d$, where $\bm{e_i}$ defines the \emph{base position} of the entity, and $\bm{b_i}$ defines its \emph{translational bump}, which translates all the entities co-occuring in a fact with $e_i$, from their base positions to their final embeddings by ``bumping'' them.
The \emph{final embedding} of an entity $e_i$ relative to a fact $r(e_{1}, \ldots , e_{n})$ is hence given by: 
\begin{align}
\label{finalemb}
{\bm{e_i^{r(e_{1}, \ldots , e_{n})}} = (\bm{e_i} - \bm{b_i}) + \sum_{1 \leq j \leq n} \bm{b_{j}}}.
\end{align}

Essentially, the entity representation is dynamic, as every entity can have a potentially different final embedding relative to a different fact. The main idea is that every entity translates the base positions of other entities co-appearing in a fact, that is, for a fact $r(e_1, e_2)$, $\bm{b_1}$ and $\bm{b_2}$ translate $\bm{e_2}$ and $\bm{e_1}$ respectively, to compute their final embeddings.

In BoxE, every relation $r$ is represented by $n$ hyper-rectangles, i.e., boxes, $\bm{r^{(1)}}, \ldots, \bm{r^{(n)}} \in \mathbb{R}^d$, where $n$ is the arity of $r$. 
Intuitively, this representation defines $n$ \emph{regions} in $\mathbb{R}^d$, one per arity position, such that a fact $r(e_1, ..., e_n)$ holds when the final embeddings of $e_1, ..., e_n$ each appear in their corresponding position box, creating a \emph{class} abstraction for the sets of all entities appearing at every arity position.
For the special case of unary relations (i.e., classes), the definition given in Eq. \ref{finalemb} implies no translational bumps, and thus the base  position of an entity is its final embedding.

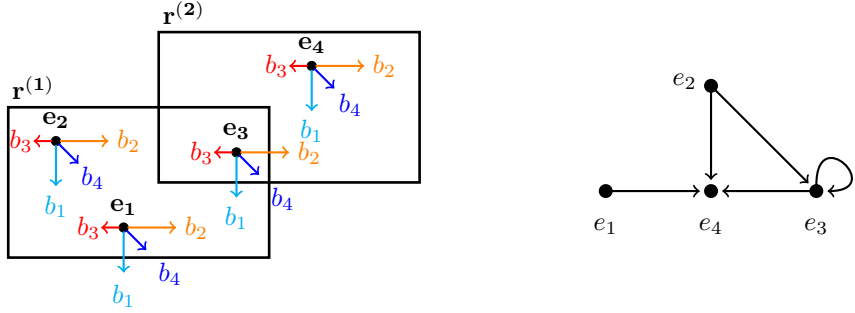
\begin{figure}[t!]
	\centering
	\begin{subfigure}{0.55\linewidth}	
	\begin{tikzpicture}[scale=1,every node/.style={minimum size=1cm},on grid]
	\tikzstyle{box} = [draw=black, line width=1pt, rectangle, scale=2]
	\tikzstyle{entity} = [fill=black, draw=black, line width=1pt,circle, scale=0.1]
	
	\node[box,label={[xshift=-1.4cm, yshift=-8pt]$\mathbf{r^{(1)}}$}] (r11) {\hspace*{1.5cm}};
	\node[box, above right=1cm and 2cm of r11,label={[xshift=-1.4cm, yshift=-8pt]$\mathbf{r^{(2)}}$}] (r12) {\hspace*{1.5cm}};
	
	\node[entity, below left=0.6cm and 0.2cm of r11,label={[xshift=0pt, yshift=-8pt]$\mathbf{e_1}$}] (e1) {};
	\node[entity, above left=0.55cm and 1.1cm of r11,label={[xshift=0pt, yshift=-8pt]$\mathbf{e_2}$}] (e2) {};
	\node[entity, below left=0.6cm and 0.7cm of r12,label={[xshift=0pt, yshift=-8pt]$\mathbf{e_3}$}] (e3) {};
	\node[entity, above right=0.55cm and 0.3cm of r12,label={[xshift=0pt, yshift=-8pt]$\mathbf{e_4}$}] (e4) {};
	
	\draw[->, cyan,thick](e1) to ($(e1)+(0,-0.6)$) node[below=-6pt] {$b_1$};
	\draw[->, orange,thick](e1) to ($(e1)+(0.7,0)$) node[right=-7pt] {$b_2$};	
	\draw[->, red,thick](e1) to ($(e1)+(-0.3,0)$) node[left=-10pt] {$b_3$};
	\draw[->, blue,thick](e1) to ($(e1)+(0.3,-0.3)$) node[below right=-8pt] {$b_4$};
	\draw[->, cyan,thick](e2) to ($(e2)+(0,-0.6)$) node[below=-7pt] {$b_1$};
	\draw[->, orange,thick](e2) to ($(e2)+(0.7,0)$) node[right=-7pt] {$b_2$};		
	\draw[->, red,thick](e2) to ($(e2)+(-0.3,0)$) node[left=-10pt] {$b_3$};
	\draw[->, blue, thick](e2) to ($(e2)+(0.3,-0.3)$) node[below right=-13pt] {$b_4$};	
	\draw[->, cyan,thick](e3) to ($(e3)+(0,-0.6)$) node[below=-7pt] {$b_1$};
	\draw[->, orange,thick](e3) to ($(e3)+(0.7,0)$) node[right=-7pt] {$b_2$};
	\draw[->, red,thick](e3) to ($(e3)+(-0.3,0)$) node[left=-10pt] {$b_3$};
	\draw[->, blue, thick](e3) to ($(e3)+(0.3,-0.3)$) node[below right=-8pt] {$b_4$};	
	\draw[->,cyan,thick](e4) to ($(e4)+(0,-0.6)$) node[below=-7pt] {$b_1$};
	\draw[->,orange,thick](e4) to ($(e4)+(0.7,0)$) node[right=-7pt] {$b_2$};	
	\draw[->,red,thick](e4) to ($(e4)+(-0.3,0)$) node[left=-10pt] {$b_3$};
	\draw[->, blue, thick](e4) to ($(e4)+(0.3,-0.3)$) node[below right=-12pt] {$b_4$};	
	\end{tikzpicture}
	\end{subfigure}
	\begin{subfigure}{0.4\linewidth}	
	\begin{tikzpicture}[scale=0.7,every node/.style={minimum size=1cm},on grid]
	
	\tikzstyle{vertex} = [fill=black, draw=black, line width=1pt,circle, scale=0.15]
	\tikzstyle{edge} = [->, shorten >=2pt, thick]
	\tikzset{every loop/.style={min distance=10mm,in=0,out=90,looseness=20}}

	\node[vertex,label={[xshift=0pt, yshift=-30pt]$e_1$}] (e1) at  (0,0) {};
	\node[vertex,label={[xshift=-10pt, yshift=-15pt]$e_2$}] (e2) at  (2,2) {};
	\node[vertex, label={[xshift=0pt, yshift=-30pt]$e_4$}] (e4) at  (2,0) {};
	\node[vertex,label={[xshift=0pt, yshift=-30pt]$e_3$}] (e3) at  (4,0) {};
	\draw[edge] (e1) to (e4);
	\draw[edge] (e2) to (e4);
	\draw[edge] (e2) to (e3);	
 	\draw[edge] (e3) [loop above] node {} to (e3);
	\draw[edge] (e3) to  (e4);
	\end{tikzpicture}
	\end{subfigure}
	\caption{A sample BoxE model is shown on the left for $d=2$. The binary relation $r$  is encoded via the box embeddings $\mathbf{r^{(1)}}$ and $\mathbf{r^{(2)}}$. Every entity $e_i$ has an embedding $\mathbf{e_i}$, and defines a bump on other entities, as shown with distinct colors. This model induces the KG on $r$, shown on the right.}
	\label{fig:BoxEBumps}
\end{figure}
\begin{example}
{\rm 
Consider an example over a single binary relation $r$ and the entities $e_1, e_2, e_3, e_4$.
A~BoxE model is given on the left in  \Cref{fig:BoxEBumps}, for $d=2$, where every entity is represented as a point, and the binary relation $r$ is represented with two boxes $\bm{r^{(1)}}$ and $\bm{r^{(2)}}$.
Every entity is translated by the bump vectors of all other entities. For example, $r(e_1,e_4)$ is a true fact in the model (e.g., to be ranked high), since (i)  $\bm{e_1}^{\bm{r(e_1,e_4)}}=(\bm{e_1}+\bm{b_4})$ is a point in $\bm{r^{(1)}}$ ($\bm{e_1}$ appears in the head box), and (ii) $\bm{e_4}^{\bm{r(e_1,e_4)}}=(\bm{e_4}+\bm{b_1})$ is a point in $\bm{r^{(2)}}$ ($\bm{e_4}$ appears in the tail box). Similarly, $r(e_3, e_3)$ is a true fact in the model, as  ${\bm{e_3}^{\bm{r(e_3,e_3)}}=(\bm{e_3}+\bm{b_3})}$, which is a point in $\bm{r^{(1)}}$ and $\bm{r^{(2)}}$, i.e., the entity is reflexive in $r$. The model encodes all (and only) the facts from the KG, shown on the right in \Cref{fig:BoxEBumps}. \hfill$\rule{0.5em}{0.5em}$

}
\end{example}

Translational bumps are very powerful, as they allow us to model complex interactions across entities in an effective manner. Observe that for the sample KG, there are $4^2$ potential facts that can hold, and therefore $2^{16}$ possible configurations. Nonetheless, they can all be compactly captured by choosing appropriate translational bumps to force entity embeddings in or out of the respective relation boxes as needed. Indeed, we later formally show that such a configuration can always be found for any KB, given sufficiently many dimensions, proving full expressiveness of the model.

\textbf{Scoring function.} 
In the above example, we identified facts that ideally need to be ranked higher by our scoring function,  to reflect the model properties adequately.
To this end, we first define a distance function for evaluating entity positions relative to the box positions.
The idea is to define a function that grows slowly if a point is in the box (relative to the center of the box), but grows rapidly if the point is outside of the box, so as to drive points more effectively into their target boxes and ensure they are minimally changed, and can remain there once inside.

Formally, let us denote by $\bm{l^{(i)}}, \bm{u^{(i)}} \in \mathbb{R}^d$ the \emph{lower} and \emph{upper} boundaries of a box $\bm{r^{(i)}}$, respectively, by ${\bm{c}^{(i)} = \sfrac{(\bm{l^{(i)}} + \bm{u^{(i)}})}{2}}$ its center, and by ${\bm{w}^{(i)} = \bm{u^{(i)}} - \bm{l^{(i)}} + 1}$ its width incremented by 1. We say that a point $\bm{e_i}$ is inside a box $\bm{r^{(i)}}$, denoted $\bm{e_i} \in \bm{r^{(i)}}$, if $\bm{l^{(i)}} \leq \bm{e_i} \leq \bm{u^{(i)}}$. Furthermore, we denote the element-wise multiplication, division, and  inversion operations by $\circ, \oslash$ and $^{\circ-1}$ respectively. Then, the \emph{distance function} for the given entity embeddings relative to a given target box is defined piece-wise over two cases, as follows:
\[
\dist(\bm{e_i^{r(e_1, ..., e_n)}}, \bm{r^{(i)}}) = 
\begin{cases} 
\mid \bm{e_i^{r(e_1, ..., e_n)}} - \bm{c^{(i)}}\mid \oslash~\bm{w^{(i)}} & \text{if} \, \bm{e_i} \in \bm{r^{(i)}}, \\ 
\mid\bm{e_i^{r(e_1, ..., e_n)}} - \bm{c^{(i)}}\mid \circ~\bm{w^{(i)}} -  \kappa & \text{otherwise,}
\end{cases}
\]%
where $\kappa = 0.5\circ(\bm{w^{(i)}}-1)\circ(\bm{w^{(i)}}-\bm{w^{(i)^{\circ-1}}})$, is a width-dependent factor.

\begin{wrapfigure}[16]{r}{0.35\textwidth}
\centering
\begin{tikzpicture}[scale=0.67,
    declare function={
    width0(\x)= (\x<=0) * (-\x) + (\x>0) * (\x));
  },
  declare function={
    width2(\x)= (\x<=-1) * (-3*\x -2.66666667)   +
     and(\x>-1, \x<=0) * (-0.333333*\x)     +
     and(\x>0,  \x<=1) * (0.333333*\x) +
                (\x>1) * (3*\x - 2.66666667);
  }, declare function={
    width4(\x)= (\x<=-2) * (-5*\x -9.6)   +
     and(\x>-2, \x<=0) * (-0.2*\x)     +
     and(\x>0,  \x<=2) * (0.2*\x) +
                (\x>2) * (5*\x - 9.6);
  }
]
\definecolor{color0}{rgb}{0.12156862745098,0.466666666666667,0.705882352941177}
\definecolor{color1}{rgb}{1,0.498039215686275,0.0549019607843137}
\definecolor{color2}{rgb}{0.172549019607843,0.627450980392157,0.172549019607843}
\definecolor{color3}{rgb}{0.83921568627451,0.152941176470588,0.156862745098039}
\begin{axis}[
legend cell align={left},
legend entries={{$w^{(i)}=1$},{$w^{(i)}=3$},{$w^{(i)}=5$}},
legend style={at={(0.13,0.99)}, line width=0.6mm, anchor=north west, draw=white!80.0!black},
tick align=outside,
  axis x line=middle, axis y line=middle,
  ymin=-0.1, ymax=10, ytick={0,...,9}, ylabel=$\dist$,
  xmin=-4, xmax=4, xtick={-4,...,4}, xlabel=$\bm{e_i^{r(e_1,...,e_n)}} - \bm{c^{(i)}}$,
  x label style={at={(axis description cs:0.81,-0.05)},anchor=north, font=\large},
  y label style={at={(axis description cs:0.58,.9)},rotate=90,anchor=south},
]
\addplot[color1, domain=-4:4, line width=0.7mm]{width0(x)};
\addplot[color0, domain=-4:4, line width=0.7mm]{width2(x)};
\addplot[color3, domain=-4:4, line width=0.7mm]{width4(x)};
\end{axis}
\end{tikzpicture}
\caption{The $\dist$ function for width ${\bm{w^{(i)}}=1,3,5}$.}
\label{fig:boxdist}
\end{wrapfigure}
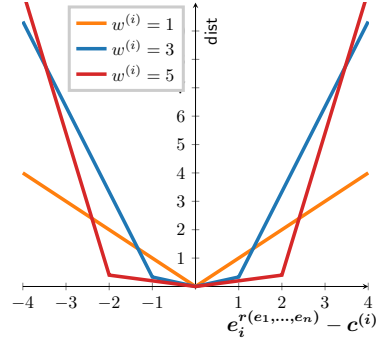

In both cases, $\dist$ factors in the size of the target box in its computation. In the first case, where the point is in its target box, distance inversely correlates with box size, to maintain low distance inside large boxes and provide a gradient to keep points inside. In the second case, box size linearly correlates with distance, to penalize points outside larger boxes more severely. Finally, $\kappa$ is subtracted to preserve function continuity.

Plots for $\dist$ for one-dimensional $\bm{w^{(i)}}$ are shown in \Cref{fig:boxdist}. Observe that, when $\bm{w^{(i)}}=1$, $\bm{r^{(i)}}$ is point-shaped, and $\dist$ reduces to standard $\text{L}^1$ distance. Conversely, as $\bm{w^{(i)}}$ increases, $\dist$ gives lower values (and gradients) to the region inside the box, and severely punishes points outside. 
This function thus achieves three objectives. First, it treats points inside the box preferentially to points outside the box, unlike  standard distance, which is agnostic to boxes. Second, it ensures that outside points receive high gradient through which they can more easily reach their target box, or escape it for negative samples. Third, it gives weight to the size of a box in distance computation, to yield a more comprehensive scoring mechanism.

Finally, we 
define the scoring function as the sum of the L-$x$ norms of $\dist$ across all $n$ entities and relation boxes, i.e.: 
\[
\score(r(e_1, ..., e_n)) = \sum_{i=1}^{n} \left\Vert \dist(\bm{e_i^{r(e_1, ..., e_n)}}, \bm{r^{(i)}})  \right\Vert_x.
\]

\section{Model Properties}
\label{sec:BoxEProps}
We analyze the representation power and inductive capacity of BoxE and show that BoxE is fully expressive, and can capture a rich language combining multiple inference patterns. We additionally show that BoxE can lucidly \emph{incorporate} a given set of logical rules from a sublanguage of this language, i.e., rule injection. Finally, we analyze the complexity of BoxE in the appendix, and prove that it runs in time $O(nd)$ and space $O((|\Ebf| + n|\Rbf|)d)$, where $n$ is the maximal relation arity. 

\subsection{Full expressiveness}
We prove that BoxE is fully expressive with $d = |\Ebf|^{n-1}|\Rbf|$ dimensions. For KGs, this result implies $d = |\Ebf||\Rbf|$, so BoxE is fully expressive over KGs with dimensionality \emph{linear} in $|\Ebf|$. The proof uses translational bumps to make an arbitrary true fact $F$ false, while preserving the correctness of other facts. This result requires a careful technical construction, which (i) pushes a single entity representation within $F$ outside its corresponding relation box at a specific dimension, and (ii)~modifies all other model embeddings to prevent a change in the truth value of any other fact.

\begin{theorem}
\label{thm:fullexp}
BoxE is a fully expressive model with the embedding dimensionality $d$ of entities, bumps, and relations set to $d = |\Ebf|^{n-1}|\Rbf|$, where $n>1$ is the maximal arity of the relations in $\Rbf$.
\end{theorem}
We note that this result makes BoxE the first translation-based model that is fully expressive.

\begin{table}[t]
	\centering
	\caption{Inference patterns/generalized inference patterns captured by selected KBC models. TuckER coincides with ComplEx, so is omitted from the table.}
	\begin{tabular}{lc@{\hskip 8pt}c@{\hskip 8pt}c@{\hskip 8pt}c@{\hskip 8pt}cH}
		\toprule
		{\textbf{Inference pattern}} & BoxE & TransE  & RotatE  & DistMult & ComplEx & Tucker\\
		\cmidrule(l){1-7}
		 Symmetry: $r_1(x,y) \Rightarrow r_1(y,x)$ &
		 {\cmark/\cmark} &
		 {\xmark/\xmark} &
		 {\cmark/\cmark} &
		 {\cmark/\cmark} &
		 {\cmark/\cmark} &
		 {\cmark/\cmark}  \\
		 Anti-symmetry: $r_1(x,y)  \Rightarrow \neg r_1(y,x)$ & 
		 {\cmark/\cmark} &  
		 {\cmark/\cmark} &  
		 {\cmark/\cmark} &
		 {\xmark/\xmark} &
		 {\cmark/\cmark} &
		 {\cmark/\cmark}   \\
		 Inversion: $r_1(x,y) \Leftrightarrow r_2(y,x)$  & 
		 {\cmark/\cmark} &  
		 {\cmark/\xmark} &  
		 {\cmark/\cmark} &
		 {\xmark/\xmark} &
		 {\cmark/\cmark} &
		 {\cmark/\cmark}  \\
		 Composition: $r_1(x,y) \land r_2(y,z) \Rightarrow r_3(x,z)$  &  {\xmark/\xmark} & 
		 {\cmark/\xmark} & 
		 {\cmark/\xmark} &  
		 {\xmark/\xmark} &  
		 {\xmark/\xmark} &  
		 {\xmark/\xmark} \\
		Hierarchy: $r_1(x,y) \Rightarrow r_2(x,y)$  & 
		{\cmark/\cmark} &  
		{\xmark/\xmark} &   
		{\xmark/\xmark} &   
		{\cmark/\xmark} &   
		{\cmark/\xmark} &   
		{\cmark/\xmark}  \\
		Intersection: $r_1(x,y) \land r_2(x,y) \Rightarrow r_3(x,y)$   & {\cmark/\cmark} & 
		{\cmark/\xmark} & 
		{\cmark/\xmark} &  
		{\xmark/\xmark} &  
		{\xmark/\xmark} &  
		{\xmark/\xmark}  \\
		Mutual exclusion: $r_1(x,y)   \land  r_2(x,y) \Rightarrow \bot$   & 
		{\cmark/\cmark} & 
		{\cmark/\cmark} &  
		{\cmark/\cmark} &  
		{\cmark/\xmark} &  
		{\cmark/\xmark} &  
		{\cmark/\xmark} \\
		\bottomrule
	\end{tabular}
	\label{tab:infPat}
\end{table}

\subsection{Inference patterns and generalizations}  
We study the inductive capacity of BoxE in terms of common inference patterns appearing in the  KGC literature, and compare it with earlier models. A comparison of BoxE against these models with respect to capturing prominent inference patterns is shown in \Cref{tab:infPat}.

A model \emph{captures} an inference pattern if it admits a set of parameters \emph{exactly} and \emph{exclusively} satisfying the pattern. This is the standard definition of an inference pattern in the literature \cite{RotatE-ICLR19}.
For example, TransE can capture composition~\cite{TransE-NIPS13,RotatE-ICLR19}, but cannot capture hierarchy, as for TransE,  ${r_1(x,y) \Rightarrow r_2(x,y)}$ holds only if $r_1 = r_2$, and thus $r_2(x,y) \Rightarrow r_1(x,y)$, leading to loss of generality. However, this definition only addresses \emph{single} applications of an inference pattern, which raises the question: can KBC models capture \emph{multiple, distinct} instances of the \emph{same} inference pattern jointly? 

Capturing multiple inference patterns jointly is significantly more challenging. Indeed, TransE can capture ${r_1(x,y) \land r_2(y,z) \Rightarrow r_3(x,z)}$ and ${r_1(x,y) \land r_4(y,z) \Rightarrow r_3(x,z)}$ independently, but jointly capturing these compositions incorrectly forces ${r_2 \sim r_4}$. Similarly, bilinear models can capture the hierarchy rules ${r_1(x,y) \Rightarrow r_3(x,y)}$ and ${r_2(x,y) \Rightarrow r_3(x,y)}$ separately, but jointly capturing them incorrectly imposes either ${r_1(x,y) \Rightarrow r_2(x,y)}$ or ${r_2(x,y) \Rightarrow r_1(x,y)}$~\cite{Gutirrez18}. These examples are clearly not edge cases, and highlight severe limitations in how the inductive capacity of KBC models is analyzed. Therefore, we propose and study \emph{generalized inference patterns}.

\begin{definition}
\label{def:gip}
A \emph{rule} is in one of the forms given in \Cref{tab:infPat}, where $r_1 \neq r_2  \neq r_3 \in \Rbf$.
To distinguish between types of rules, we write \emph{$\sigma$ rule}, where $\sigma \in \{\text{symmetry, \ldots, mutual exclusion}\}$.
A~\emph{generalized $\sigma$} pattern is a finite set of $\sigma$ rules over $\Rbf$. 
\end{definition}

As before, a model \emph{captures} a generalized inference pattern if the model admits a set of parameters, exactly and exclusively satisfying the generalized pattern. Our results for BoxE and all relevant models are summarized in \Cref{tab:infPat}, and proven in the following theorem. %
\begin{theorem}
\label{thm:genPat}
All the results given in \Cref{tab:infPat} for BoxE and other models hold.
\end{theorem}

Intuitively, BoxE captures all these generalized inference patterns through box configurations. %
For instance, BoxE captures (generalized) symmetry by setting the 2 boxes for a relation $r$ to be equal, and captures (generalized) inverse relations $r_1$ and $r_{2}$ by setting $\bm{r_1^{(1)}} = \bm{r_{2}^{(2)}}$ and $\bm{r_1^{(2)}} = \bm{r_{2}^{(1)}}$. Hierarchies are captured through box subsumption, i.e., $\bm{r_1^{(1)}}$ and $\bm{r_1^{(2)}}$ contained in $\bm{r_{2}^{(1)}}$ and $\bm{r_{2}^{(2)}}$ respectively, and this extends to intersection in the usual sense. Finally, anti-symmetry and mutual exclusion, are captured through disjointness between relation boxes.

Generalized inference patterns are necessary to establish a more complete understanding of model inductive capacity, and, in this respect,  %
our results show that BoxE goes well beyond any other model. However, generalized patterns are not sufficient. Indeed, different types of inference rules can appear \emph{jointly} in practical applications, so KBC models must be able to jointly capture them. This is not the case for existing models. For instance, RotatE can capture composition and generalized symmetry, but to capture a single composition rule such as $\mathsf{cousins}(x,y)  \land \mathsf{hasChild}(y,z)\Rightarrow \mathsf{relatives}(x,z)$, 
where $\mathsf{relatives}$ and $\mathsf{cousins}$ are \emph{symmetric} relations, the model forces $\mathsf{hasChild}$ to be symmetric as well, i.e., $\mathsf{hasChild}(x,y) \Rightarrow \mathsf{hasChild}(y,x)$, which is clearly absurd.
Therefore, we also evaluate model inductive capacity relative to more general \emph{rule languages}~\cite{Gutirrez18}. We define a rule language as the \emph{union} of different types of rules. Thus,  generalized inference patterns are trivial rule languages allowing only one type of rule. 
BoxE can capture rules from a rich language, as stated next.

\begin{theorem}
\label{thm:InfPat} 
Let $\Lmc$ be the rule language that is the union of inverse, symmetry, hierarchy, intersection, mutual exclusion, and anti-symmetry rules.
BoxE can capture any finite set of consistent rules from the rule language $\Lmc$.
\end{theorem}

This result captures generalized inference patterns for BoxE as a special case. Such a result is implausible for other KBC models, given their limitations in capturing generalized inference patterns, and we are unaware of any analogous result in KBC. The only related result is for ontology embeddings, and for quasi-chained rules~\cite{Gutirrez18}, but this result merely offers region structures enabling capturing a set of rules, without providing any viable model or means of doing so. 

The strong inductive capacity of BoxE is advantageous from an interpretability perspective, as all the rules that BoxE can jointly capture can be simply ``read'' from the corresponding box configuration. Indeed, BoxE embeddings allow for rich rule extraction, and enable an informed understanding of what the model learns, and how it reaches its scores. This is a very useful consequence of inductive capacity, as better rule capturing directly translates into superior model interpretability. Finally, BoxE can seamlessly and naturally represent \emph{entity type} information, e.g., $\mathsf{country(UK)}$ by modeling types as \emph{unary} relations. In this setting, translational bumps are not applicable, and inference patterns deducible from classic box configurations can additionally be captured and extracted. By contrast, standard models require dedicated modifications to their parameters and scoring function \cite{XieLS16,ChangYYM14,LvHLL18} to incorporate type information. This therefore further highlights the strong inductive capacity of BoxE, and its position as a unifying model for multi-arity knowledge base completion.

\subsection{Rule injection}
We now pose a complementary question to capturing inference patterns: can a KBC model be injected with a \emph{given} set of rules such that it provably enforces them, improving its prediction performance? 
Formally, we say that a rule $\phi \Rightarrow \psi$ (resp., $\psi \Leftrightarrow \phi$) can be \emph{injected} to a model, if the model can be configured to force $\psi$ to hold whenever $\phi$ holds (resp., $\phi$ holds whenever $\psi$ holds and vice versa). 

There is a subtle difference between \emph{capturing}  and \emph{injecting} an inference pattern. Indeed, rules with negation, such a mutual exclusion, can be easily captured with any disjointness between $r_1$ and $r_2$, but enforcing such a rule leads to non-determinism. 
To illustrate, $r_1$ and $r_2$ can be disjoint between their (i) head boxes, or (ii) tail boxes, or (iii) both, and at any combination of dimensions. This non-determinism only becomes more intricate as interactions across different rules are considered. We show that the positive fragment of the rule language that can be captured by BoxE, can be injected. 
\begin{theorem}
\label{thm:ruleInj}
Let $\Lmc^+$ be the rule language that is the union of inverse, symmetry, hierarchy, and intersection rules.
BoxE can be injected with any finite set of rules from the rule language $\Lmc^+$.
\end{theorem}

Existing KGC rule injection methods (i) use rule-based training loss to inject rules \cite{demeester16,rocktaschel15}, potentially leveraging fuzzy logic \cite{guo16} and adversarial training \cite{minervini17}, but cannot provably enforce rules, or (ii) constrain embeddings explicitly \cite{ding18,rocktaschel15}, but only enforce very limited rules (e.g., inversion, linear implication). Indeed, most popular standard KGC methods fail to capture simple sets of rules \cite{Gutirrez18}. 
BoxE is a powerful model for rule injection in that it can explicitly and provably enforce such rules and incorporate a strong bias by appropriately constraining the learning space. Our study is related to the broader goal of making gradient-based optimization and learning compatible with reasoning \cite{LeCunTalk}. 

\section{Experimental Evaluation}
In this section, we evaluate BoxE on a variety of tasks, namely, KGC, higher-arity KBC, and rule injection, and report state-of-the-art results, empirically confirming the theoretical strengths of BoxE.

\subsection{Knowledge graph completion}
\label{ssec:KGCRes}
In this experiment, we run BoxE on the KGC benchmarks FB15k-237, WN18RR, and YAGO3-10, and compare it with translational models TransE \cite{TransE-NIPS13} and RotatE \cite{RotatE-ICLR19}, both with uniform and self-adversarial negative sampling \cite{RotatE-ICLR19}, and with bilinear models DistMult \cite{DistMult-ICLR15}, ComplEx \cite{ComplEx-ICML16}, and TuckER \cite{TuckER}.
We train BoxE for up to 1000 epochs, with validation checkpoints every 100 epochs and the checkpoint with highest MRR used for testing. We report the best published results on every dataset for all models, and, when unavailable, report our best computed results in italic. All results are for models with $d \leq 1000$, to maintain comparison fairness  \cite{TuckER}. We therefore exclude results by ComplEx \cite{N3Reg-ICML18} and DistMult \cite{ruffinelli2020you} using $d\geq2000$. The best results by category are presented in bold, and the best results overall are highlighted by a surrounding rectangle. ``(u)'' indicates uniform negative sampling, and ``(a)'' denotes self-adversarial sampling. Further details about experimental setup, as well as hyperparameter choices and dataset properties, can be found in the appendix.

\begin{table}[t] 
	\centering
	\caption{KGC results (MR, MRR, Hits@10) for BoxE and competing approaches on FB15k-237, WN18RR, and YAGO3-10. Other approach results are  best published, with sources cited per model.} 
	\label{tab:testSet} 
	\small\addtolength{\tabcolsep}{-1pt}
	\begin{tabular}{l@{\hskip 10pt}c@{\hskip 7pt}c@{\hskip 7pt}HH@{\hskip 7pt}c@{\hskip 10pt}c@{\hskip 7pt}c@{\hskip 7pt}HH@{\hskip 7pt}c@{\hskip 10pt}c@{\hskip 7pt}c@{\hskip 7pt}HH@{\hskip 7pt}c@{\hskip 7pt}}
		\toprule 
		 {Model} & \multicolumn{5}{c}{\textbf{FB15k-237}} & \multicolumn{5}{c}{\textbf{WN18RR}} & \multicolumn{5}{c}{\textbf{YAGO3-10}} \\
		\cmidrule(r){2-6}
		\cmidrule(r){7-11}
		\cmidrule(r){12-16}
		 & MR & MRR & H@1 & H@3 & H@10 & MR & MRR & H@1 & H@3 & H@10 & MR & MRR & H@1 & H@3 & H@10\\
		 TransE(u)  \cite{ruffinelli2020you} & - & .313 & - & - & .497 & - & .228 & - & - & .520 & \textit{-} & \textit{-} & \textit{-} & \textit{-} & \textit{-} \\
		 RotatE(u) \cite{RotatE-ICLR19} & 185 & .297 & .205 & .328 & .480 & \textit{3254} & \textbf{\textit{.470}} & \textbf{\textit{.422}} & \textbf{\textit{.488}} & \textbf{\textit{.564}} & \textbf{\textit{1116}} & \textit{.459} & \textit{.360} & \textit{.512} & \textit{.651}\\
		 BoxE(u) & \textbf{172} & \textbf{.318} & \textbf{.223} & \textbf{.351} & \textbf{.514} & \fbox{\textbf{3117}} & .442 & .398 & .461 & .523 & 1164 & \fbox{\textbf{.567}} & \fbox{\textbf{.494}} & \fbox{\textbf{.611}} & \fbox{\textbf{.699}}\\
		 \midrule 
		 TransE(a) \cite{RotatE-ICLR19} & 170 & .332 & .233 & .372 & .531 & 3390 & .223 & .013 & .401 & .529 & - & - & - & - & -\\
		 RotatE(a) \cite{RotatE-ICLR19} & 177 & \textbf{.338} & \textbf{.241} & \textbf{.375} & .533 & 3340 & \fbox{\textbf{.476}} & \textbf{.428} & \fbox{\textbf{.492}} & \fbox{\textbf{.571}} & 1767 & .495 & .402 & .550 & .670\\
		 BoxE(a) & \fbox{\textbf{163}} & .337 & .238 & .374 & \textbf{.538} & \textbf{3207} & .451 & .400 & .472 & .541 & \fbox{\textbf{1022}} &  \textbf{.560} & \textbf{.484} & \textbf{.608}  & \textbf{.691}\\
		 \midrule 
		 DistMult \cite{ruffinelli2020you,DistMult-ICLR15} & - & .343 & - & - & .531 & - & .452 & - & - & .531 & 5926 & .34 & .24 & .38 & .54\\
		 ComplEx \cite{ruffinelli2020you,DistMult-ICLR15} & - & .348 & - & - & .536 & - & \textbf{.475} & - & - & \textbf{.547} & 6351 & .36 & .26 & .40 & .55\\
		 TuckER \cite{TuckER} & - & \fbox{\textbf{.358}} & \fbox{\textbf{.266}} & \fbox{\textbf{.394}} & \fbox{\textbf{.544}} & - & .470 & \fbox{\textbf{.443}} & \textbf{.482} & .526 & \textbf{\textit{4423}} & \textbf{\textit{.529}} & \textbf{\textit{.451}} &  \textbf{\textit{.576}} & \textbf{\textit{.670}}\\
		\bottomrule
	\end{tabular}
	\label{tab:KGCResults}
\end{table}

\textbf{Results.} For every dataset and model, MR, MRR, and  Hits@10 are reported in Table \ref{tab:KGCResults}. On FB15k-237, BoxE performs best among translational models, and is competitive with TuckER, especially in Hits@10. 
Furthermore, BoxE is comfortably state-of-the-art on YAGO3-10, significantly surpassing RotatE and TuckER. This result is especially encouraging considering that YAGO3-10 is the largest of all three datasets, and involves a challenging combination of inference patterns, and many fact appearances per entity. On YAGO3-10, we also observe that BoxE successfully learns symmetric relations, and learns box sizes correlating strongly with relational properties (cf. Appendix). Strong BoxE performance on FB15k-237, which contains several composition patterns, suggests that BoxE can perform well with compositions, despite not capturing them explicitly as an inference pattern.%

On WN18RR, BoxE performs well in terms of MR, but is less competitive with RotatE in MRR. We investigated WN18RR more deeply, and identified two main factors for this. First, WN18RR primarily consists of hierarchical knowledge, which is logically flattened into deep tree-shaped compositions, such as $\mathsf{hypernym(spoon, utensil)}$.
Second, symmetry is prevalent in WN18RR, e.g., $\mathsf{derivationally\_related\_form}$ accounts for 29,715 ($\sim$34.5\%) of WN18RR facts, which, combined with compositions, also helps RotatE. Indeed, in RotatE, the composition of two symmetric relations is (incorrectly) symmetric, but this is useful for WN18RR, where 4 of the the 11 relations are symmetric. That is, the modelling limitations of RotatE become an advantage given the  setup of WN18RR, and enable it to achieve state-of-the-art performance on this dataset.
 
Overall, BoxE is competitive on all %
benchmarks %
, and is state of the art on YAGO3-10. Hence, it is a strong model for KGC on large, real-world KGs. We also evaluated the robustness of BoxE relative to dimensionality on YAGO3-10, and analyzed the resulting box configuration on this dataset from an interpretability perspective. These additional experiments can be found in the appendix.

\subsection{Higher-arity knowledge base completion}
\begin{wraptable}[14]{r}{0.52\textwidth}
	\centering
	\caption{KBC results on JF17K and FB-AUTO.} 
	\label{tab:testSetYAGO} 
	\begin{tabular}{lHcHHcHcHHc}
		\toprule 
		\multirow{2}{*}{Model} & \multicolumn{5}{c}{\textbf{JF17K}} & \multicolumn{5}{c}{\textbf{FB-AUTO}} \\
		\cmidrule(r){2-6}
		\cmidrule(r){7-11}
		 & MR & MRR & H@1 & H@3 & H@10 & MR & MRR & H@1 & H@3 & H@10\\
		 m-TransH & - & .446 & .357 & .495 & .614 & - & .728 & 0.727 & .728 & .728\\
		 m-DistMult & - & .460 & .367& .510 & .635 & - & .784 & .745 & .815 & .845\\
		 m-CP & - & .392 & .303 & .441& .560 & - & .752 & .704 & .785 & .837\\
		 HypE & - & .492 & .409 & .533 & .650 & - & .804 & .774 & .823 & .856\\
		 HSimplE & - & .472 & .375 & .523 & .649 & - & .798 & .766 & .821 & .855\\
		 BoxE(u) & \fbox{\textbf{363}} & .553 & .467 & .596 & .711 & \fbox{\textbf{110}} & .837 & .804 & .858 & .895\\
		 BoxE(a) & 372 & \fbox{\textbf{.560}} & \fbox{\textbf{.472}} & \fbox{\textbf{.604}} & \fbox{\textbf{.722}} & 122 & \fbox{\textbf{.844}} & \fbox{\textbf{.814}} & \fbox{\textbf{.863}} & \fbox{\textbf{.898}}\\
		\bottomrule
	\end{tabular}
	\label{tab:multiArity}
\end{wraptable}

In this experiment, we evaluate BoxE on datasets with \emph{higher arity}, namely the publicly available JF-17K and FB-AUTO. These datasets contain facts with arities up to 6 and 5, respectively, and include facts with \emph{different arities}, i.e., 2, 3, 4, and 5. 
We compare BoxE with the best-known reported results over the same datasets \cite{Fatemi19}. For this experiment, we set $d=200$, for fairness with other models, and perform hyperparameter tuning analogously to  \Cref{ssec:KGCRes}.

\textbf{Results.}  MRR and Hits@10 for all evaluated models are given in \Cref{tab:multiArity}. On both datasets, BoxE achieves state-of-the-art performance. This is primarily due to the natural extensibility of BoxE to non-uniform and higher arity. Indeed, BoxE defines unique boxes for every arity position, enabling a more natural representation %
of entity sets at every relation position. By contrast, all other models represent all relations with identical embedding structures, which can bottleneck the learning process, in particular when arities vary. Furthermore, the inductive capacity of BoxE also naturally extends to higher arities as a result of its structure, namely for higher-arity hierarchy, intersection, and mutual exclusion, which further improves its learning ability in this setting.

\subsection{Rule injection}
\savebox\ForestBox{
\forestset{
  my tier/.style={%
    tier/.wrap pgfmath arg={level##1}{level()},
  },
}
\scalebox{0.94}
{
\begin{forest}
  for tree={
    grow'=0,
    child anchor=west,
    parent anchor=south,
    anchor=west,
    calign=first,
    s sep+=-6pt,
    inner sep=1.5pt,
    edge path={
      \noexpand\path [draw, \forestoption{edge}]
      (!u.south west) +(7.5pt,0) |- (.child anchor)\forestoption{edge label};
    },
    before typesetting nodes={
      if n=1{
        insert before={[, phantom, my tier]},
      }{},
    },
    my tier,
    fit=band,
    before computing xy={%
      l=10pt,
    },
  }
  [
    [{$\mathsf{AgentBelongsToOrganization}$}
      [$\mathsf{TeamPlaysInLeague}$]
      [$\mathsf{PersonBelongsToOrganization}$
        [$\mathsf{AthletePlayedForSchool}$]
        [$\mathsf{AthletePlaysForTeam}$]
        [$\mathsf{AthletePlaysInLeague}$
          [$\mathsf{AthleteLedSportsTeam}$]
        ]
        [$\mathsf{CoachesInLeague}$]
        [$\mathsf{WorksFor}$
          [$\mathsf{CoachesTeam}$]
        ]
      ]
    ]
    [$\mathsf{AthleteCoach}$]
  ]
\end{forest}
}
}

\begin{wrapfigure}[13]{l}{0.37\textwidth}
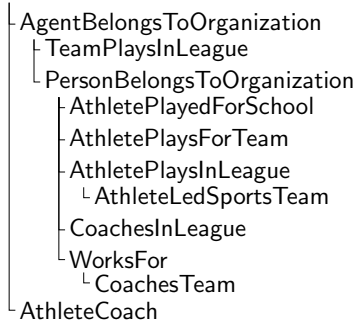

    \centering
\vspace{-0.5cm}
\usebox\ForestBox
\caption{The SportsNELL ontology.}
\label{fig:SportsNELL}
\end{wrapfigure}
In this experiment, we investigate the impact of rule injection on BoxE  performance on the SportsNELL dataset, a subset of NELL \cite{MitchellBCM18} with a known ontology, shown in \Cref{fig:SportsNELL}.
We also consider the dataset $\text{SportsNELL}^\text{C}$, which is precisely the logical closure of the SportsNELL dataset w.r.t.\ the given ontology (i.e., completion of SportsNELL under the rules). 
\begin{wraptable}[11]{r}{0.55\textwidth}
	\centering
	\caption{Rule injection experiment results on the SportsNELL full and filtered evaluation sets.} 
	\label{tab:SportsNELLRuleInj} 
	\begin{tabular}{l@{\hskip 5pt}c@{\hskip 3pt}c@{\hskip 3pt}c@{\hskip 5pt}c@{\hskip 3pt}c@{\hskip 3pt}c@{\hskip 3pt}}
		\toprule 
		Model & \multicolumn{3}{c}{Full Set} & \multicolumn{3}{c}{Filtered  Set} \\
		\cmidrule(r){2-4}\cmidrule(r){5-7}
		 & MR & MRR & H@10 & MR & MRR & H@10\\
		 BoxE & 17.4 & .577 & .780 & 19.1 & .713 & .824\\
		 BoxE+RI & \fbox{\textbf{1.74}} & \fbox{\textbf{.979}} & \fbox{\textbf{.997}} & \fbox{\textbf{5.11}} & \fbox{\textbf{.954}} & \fbox{\textbf{.984}}\\
		\bottomrule
	\end{tabular}
\end{wraptable}

We compare plain BoxE with BoxE injected with the SportsNELL ontology, denoted BoxE+RI. We train both models for 2000 epochs on a random subset (90\%) of {SportsNELL}. First, we evaluate both models on all remaining facts from  $\text{SportsNELL}^\text{C}$, which we refer to as the \emph{full evaluation set}, to measure the effect of rule injection. 
Second, we evaluate both models on a subset of the full evaluation set, only consisting of facts that are \emph{not} directly deducible via the ontology from the training set (i.e., eliminating all inferences that can be made by a rule-based approach alone). 
This subset, which we call the \emph{filtered evaluation set},  thus carefully tests the impact of rule injection on model inductive capacity.

\textbf{Results.} The results on both evaluation datasets are shown in \Cref{tab:SportsNELLRuleInj}. On the full evaluation dataset, BoxE+RI performs significantly better than BoxE. This shows that rule injection clearly improves the performance of BoxE. Importantly, this performance improvement cannot solely be attributed to the facts that can be deduced directly from the training set (with the help of the rules), as BoxE+RI performs much better than BoxE also over the filtered evaluation set. 
These experiments suggest that rule injection improves the inductive bias of BoxE, by enforcing all predictions to also conform with the given set of rules, as required. Intuitively, all predictions get amplified with the help of the rules, a very desired property, as many real-world KBs have an associated schema, or a simple ontology. 

While allowing to amplify predictions, rule injection can potentially lead to poor performance with existing metrics. Indeed, if a model mostly predicts wrong facts, these would lead to further wrong conclusions due to rule application. Hence, a low-quality prediction model can find its performance further hindered by rule injection, as false predictions create yet more false positives, thereby lowering the rank of any good predictions in evaluation. Therefore, rule injection must be complemented with models having good inductive capacity (for sparser and simpler datasets) and expressiveness (for more complex and rich datasets), such that they yield high-quality predictions in all data settings \cite{TrouillonGDB19}.

\section{Summary}
We presented BoxE, a spatio-translational model for KBC, and proved several strong results about its representational power and inductive capacity. We then empirically showed that BoxE achieves state-of-the-art performance both for KGC, and on higher-arity and different-arity KGC. Finally, we empirically validated the impact of rule injection, and showed it improves the overall inductive bias and capacity of BoxE. Overall, BoxE presents a strong theoretical backbone for KBC, combining theoretical expressiveness with strong inductive capacity and promising empirical performance. %

\section*{Acknowledgments}
This work was supported by the Alan Turing Institute under the UK EPSRC grant EP/N510129/1, the AXA Research Fund, and by the EPSRC grants EP/R013667/1 and EP/M025268/1. Ralph Abboud is funded by the Oxford-DeepMind Graduate Scholarship and the Alun Hughes Graduate Scholarship. Experiments for this work were conducted on servers provided by the Advanced Research Computing (ARC) cluster administered by the University of Oxford. %

\section*{Broader Impact}
The representation and inference of knowledge is essential for humanity, and thus any improvements in the quality and reliability of automated inference methods can significantly support endeavors in several application domains. 
This work provides a means for dealing with incomplete knowledge, and offers users to complete their knowledge bases with the help of automated machinery. The model predictions rely mostly on interpretable and explainable logical patterns, which makes it easier to analyze the model behavior.
Furthermore, this work enables safely injecting background rules when completing knowledge bases, and this safety is of great value in settings where inferred knowledge is critical (e.g., completing medical knowledge bases). This work thus also provides a logically grounded approach that improves the quality of predictions and completions in safety-critical settings. The ability of the proposed model to naturally handle more general knowledge bases (beyond knowledge graphs) could also unlock the use of knowledge base completion technologies on important knowledge bases which were previously ignored.

\bibliographystyle{plain}
\bibliography{main.bib}

\appendix
\section{Runtime and Space Complexity of BoxE}
\label{app:complexity}
\textbf{Runtime.} 
For any fact $r(e_1, \ldots, e_n)$, we can compute the entity representations $\bm{e^{r(e_1, \ldots, e_n)}}$,  in time $O(nd)$, by first computing $\sum_{1 \leq  i \leq n} \bm{b_{i}}$ in $O(nd)$, then subtracting $\bm{b_i}$ from the overall sum for every entity $e$ and finally adding the base position $\bm{e}$, resulting in $3n$ $d-$dimensional addition/subtraction operations. The distance function $\dist$ runs in $O(d)$ for every box and entity, as it involves a fixed number of $d-$dimensional operations. Thus, running $\dist$ for all $n$ positions yields a running time of $O(nd)$. Hence, BoxE scoring runs in $O(nd)$ overall. This implies that BoxE scales linearly with the arity of the relations in a KB, and thus can be applied to this setting with minimal computational overhead. Assuming that $n$ is bounded, as is the case for KGs, BoxE runs in \emph{linear time} with respect to dimensionality $d$.

\textbf{Space complexity.} In terms of space complexity, BoxE stores 2 $d-$dimensional vectors per entity $e$, namely its base position $\bm{e}$ and bump $\bm{b}$, and stores 2 $d-$dimensional vectors per box, denoting its lower and upper corners. Hence, for a KB with $|\Ebf|$ entities and $|\Rbf|$ relations with arity $n$, BoxE requires $(|\Ebf| + n|\Rbf|)d$ parameters.

\section{Proof of Theorem \ref{thm:fullexp} (Full Expressiveness)}
\label{app:fullExp}
We first prove the result for knowledge graphs, and then show how this can be lifted to arbitrary knowledge bases with higher-arity relations.

The result is shown by induction. We start with a base case where the KG $G$ contains all facts from  the universe as true facts, and subsequently prove in the induction step that a BoxE model with $d = |\Ebf||\Rbf|$ can make any arbitrary fact in $G$ false without affecting the correctness of other facts. In this induction step, facts are made false by pushing the representation of a single entity in the fact outside its corresponding relation box at a specific dimension, and modifying the remaining embeddings in the model to prevent a change in the truth value of any other fact.

Let us assume without loss of generality that all relations and entities are indexed. Specifically, we consider relations $r_i \in \Rbf$, and entities $e_j \in \Ebf$, where $0\leq i \leq |\Rbf|-1$, and $0\leq j \leq |\Ebf|-1$.
We consider $d$-dimensional embedding vectors $\bm{v}$ with $d = |\Ebf||\Rbf|$, and write $\bm{v}(i,j)$ to refer to the vector index $i|\Ebf|+j$.
Intuitively, in our construction, the sequence of indices $\bm{v}(i,0), \ldots, \bm{v}(i,|\Ebf|-1)$  corresponds to 
a ``chunk'' reserved for the relation $r_i$.

\textbf{Base case: } We initialize the KG $G$ as the whole universe, i.e., the set of all possible facts over a given vocabulary. BoxE can trivially express $G$, by simply setting all entity and bump vectors to $0$, and all boxes as the unit box centered at $0$.

\textbf{Induction step: } In this step, we consider a true fact $r_i(e_j, e_k)$, and make this fact false without affecting the remainder of G. This can be done as follows:
\begin{enumerate}[Step 1.]
    \item Increment $\bm{b_j}(i,k)$ by a value $C$, such that: 
    \[ 
    \bm{e_k}(i,k) + \bm{b_j}(i,k) + C > \bm{u_{i}^{(2)}}(i,k).
    \] 
    \item Decrement all entity embeddings except that of $e_k$ by $C$ at dimension $(i,k)$:
    \[\forall ~k' \neq k, \bm{e_{k'}}(i,k) = \bm{e_{k'}}(i,k) - C.\] 
    \item For the relation $r_i$, grow the head box by $C$ at dimension $(i,k)$ both upwards and downwards, and grow the tail box downwards by $C$ in this dimension: 
    \begin{align*}
         &\bm{l_{i}^{(1)}}(i,k) =\bm{l_{i}^{(1)}}(i,k) - C, \\ &\bm{u_{i}^{(1)}}(i,k) =\bm{u_{i}^{(1)}}(i,k) + C, \\ &\bm{l_{i}^{(2)}}(i,k) =\bm{l_{i}^{(2)}}(i,k) - C.
    \end{align*}
    \item For all other relations $r_x \in \Rbf, x \neq i$, grow all boxes by $C$ at dimension $(i,k)$ in both directions, that is, for $ \beta \in \{1,2\}$: 
    \begin{align*}
           &\bm{l_{x}^{(\beta)}}(i,k) =\bm{l_{x}^{(\beta)}}(i,k) - C,\\ &\bm{u_{x}^{(\beta)}}(i,k) =\bm{u_{x}^{(\beta)}}(i,k) + C.
    \end{align*}
\end{enumerate}

Observe first that Step 1 makes $r_i(e_j, e_k)$ false, by pushing $\bm{e_k^{r_i(e_j, e_k)}}$ outside of $\bm{r_{i}^{(2)}}$ at dimension $(i,k)$ from above. This flips the truth value of $r_i(e_j, e_k)$, as required.

We now show that the results of Steps 1 \& 2, combined with the changes to relation boxes made in Steps 3 \& 4, which affect facts involving $r_i$ and other relations respectively, preserve the correctness/falsehood of all facts other than $r_i(e_j,e_k)$. To this end, we consider any possible fact $F = r_{i'}(e_{j'}, e_{k'})$ from the KG, and analyze the effect of the induction step at the head and tail of the fact. We need to consider the following cases:
\begin{enumerate}[{Case} 1.]
    \item \textbf{The fact $F$ is true:} To verify that $F = r_{i'}(e_{j'}, e_{k'})$ remains true after the inductive step, we analyze both the head entity $\bm{e_{j'}^F}$ and the tail entity $\bm{e_{k'}^F}$. 
    \begin{enumerate}
        \item \textbf{Head entity:} Observe that (i) $\bm{e_{j'}^F}$ can change by at most $C$ following Steps 1 \& 2, and 
        (ii) all relation head boxes are grown by $C$ in both directions in Steps~3 \& 4. These together imply that $\bm{e_{j'}^F} \in \bm{r^{(1)}}$ is guaranteed to hold provided that it was true before the induction step.
        \item \textbf{Tail entity:} If $e_{k'} \neq e_k$, then $\bm{e_{k'}^F}$ is not changed if $e_{j'} = e_j$, and decremented by $C$ at dimension $(i,k)$ otherwise. Hence, the changes to both $\bm{r_{i}^{(2)}}$ and $\bm{r_{x}^{(2)}}, x \neq i$ in Steps~3 \& 4 are sufficient to maintain $\bm{e_{k'}} \in \bm{r^{(2)}}$. 
        Conversely, if $e_{k'} = e_k$, then $\bm{e_t^F}$ is unchanged when $e_{j'} \neq e_j$, and thus $\bm{e_{k'}} \in \bm{r^{(2)}}$ still holds. Otherwise, when $e_{j'} = e_j$, $\bm{e_{k'}^F}$ is incremented by $C$, which, for $r = r_i$, makes $F$ false, as required, and for $r \neq r_i$, still keeps $\bm{e_{k'}} \in \bm{r^{(2)}}$, as all other tail boxes are grown upwards by C. 
    \end{enumerate}
    Hence, for any true fact in $G$, except the fact $r_i(e_j,e_k)$, we conclude that $\bm{e_{j'}} \in \bm{r^{(1)}}$ and $\bm{e_{k'}} \in \bm{r^{(2)}}$ continues to hold after the induction step, as required. 
    \item \textbf{The fact $F$ is false:}  To verify that $F = r_{i'}(e_{j'}, e_{k'})$ remains false, after the inductive step, we again consider the head and tail entities.
    \begin{enumerate}
        \item \textbf{Head entity:} By construction, all false facts $r_{i'}(e_{j'}, e_{k'})$  satisfy the inequality \[{\bm{e_{k'}^{r_{i'}(e_{j'}, e_{k'})}}(i',k') > \bm{u_{i'}^{(2)}}(i',k')},\] 
        and any changes to $\bm{e_{j'}^F}$ do not affect this inequality.
        \item \textbf{Tail entity:} If $e_{k'} \neq e_k$, then $F$ verifies $\bm{e^F}(i',k') > \bm{u_{i'}^{(2)}}(i',k')$, where $k' \neq k$. This inequality continues to hold regardless of the changes to $\bm{e_{k'}^F}(i,k)$. Otherwise, if $e_{k'} = e_k$, and $r_{i'} = r_i$, then $e_{j'} \neq e_j$, as $F$ is initially false, and $r_i(e_j,e_k)$ is initially true.  Furthermore, since $e_{j'} \neq e_j$, $\bm{e_{k'}^{F}}$ is unchanged, which maintains the falsehood inequality. Finally, if $r_{i'} \neq r_i$, then the falsehood inequality for $F$ holds at a dimension different than $(i,k)$. Therefore, none of the changes in the induction step affect this inequality.
    \end{enumerate}
    Hence, all false facts in $G$ remain false after the induction step, as required. 
\end{enumerate}

Thus, the induction step can make any true fact $r_i(e_j, e_k)$ in G false in a BoxE model with $d = |\Ebf||\Rbf|$ without affecting the remainder of the facts in G. Hence, all fact configurations are possible and expressible by such a BoxE model, and this model is fully expressive, as required.

This proof can be generalized to higher-arity knowledge bases. Indeed, for a maximum arity of $n$, a dimensionality $d = |\Ebf|^{n-1}|\Rbf|$ is needed for full expressiveness. All proof steps shown above would remain the same, except that (i) we define a higher-arity indexing function $(\theta_1, \theta_2, ..., \theta_K)$, which refers to vector index $\sum_{a=1}^n |\Ebf|^{n-a} \theta_i$, and (ii) grow boxes for $r_i$ and all other $r_x$ at positions 3 and onwards in both Steps 3 and 4, in addition to position 1, by $C$ in both directions (while the changes at position 2 remain the same). 

Finally, we note that the proof can be trivially extended to knowledge bases with non-uniform arities (i.e., KBs containing relations with different arities) by introducing extra parameters to relations of lower arity, and setting the correctness of the $n-$arity facts solely based on the original facts. Hence, BoxE is a fully expressive model for general KBs containing both \emph{distinct} and \emph{large} relation arities.%

\section{Proof of \Cref{thm:genPat} (Inference Patterns and Generalizations)}
\label{app:infPat}

We start by more explicitly reformulating \Cref{def:gip} in the main body of the paper.
\begin{definition} Generalized inference patterns are defined as follows:
\begin{itemize}
\item A \emph{symmetry rule} is of the form $r_1(x,y) \Rightarrow r_1(y,x)$, where $r_1 \in \Rbf$. A \emph{generalized symmetry} pattern is a finite set of symmetric rules over $\Rbf$.

\item An \emph{anti-symmetry rule} is of the form $r_1(x,y) \Rightarrow \neg r_1(y,x)$, where $r_1 \in \Rbf$. A \emph{generalized anti-symmetry} pattern is a finite set of anti-symmetric rules over $\Rbf$.

\item An \emph{inversion rule} is of the form $r_1(x,y) \Leftrightarrow r_2(y,x)$, where $r_1 \neq r_2 \in \Rbf$. A \emph{generalized inversion} pattern is a finite set of inverse rules over $\Rbf$.

\item A \emph{composition rule} is of the form $r_1(x,y) \land r_2(y,z) \Rightarrow r_3(x,z)$, where $r_1\neq r_2 \neq r_3 \in \Rbf$.  A \emph{generalized composition} pattern is a finite set of composition rules over  $\Rbf$.

\item A \emph{hierarchy rule} is of the form $r_1(x,y) \Rightarrow r_2(x,y)$,where $r_1 \neq r_2 \in \Rbf$. A \emph{generalized hierarchy} pattern is a finite set of hierarchy rules over $\Rbf$.

\item An \emph{intersection rule} is of the form $r_1(x,y) \land r_2(x,y) \Rightarrow r_3(x,y)$,  where $r_1 \neq r_2 \neq r_3 \in \Rbf$.  A \emph{generalized intersection} pattern is a finite set of intersection rules over $\Rbf$.

\item A \emph{mutual exclusion rule} is of the form $r_1(x,y) \land  r_2(x,y) \Rightarrow  \bot$,  where $r_1 \neq r_2 \in \Rbf$.  A \emph{generalized mutual exclusion} pattern is a finite set of mutually exclusive rules over $\Rbf$.
\end{itemize} 
Every generalized inference pattern defines a trivial rule language, consisting of a single type of rule.
We define more general rule languages, by taking the union of different types of rules.
A rule language $\Lmc$ is defined in terms of the types of rules that are allowed in the language.
\end{definition}

\textbf{Remark.} The requirement for setting the relations to be distinct is due to the existing conventions in the literature. This may appear somewhat unintuitive, but it is required to study the rules in isolation, i.e., a composition rule without this requirement can express transitivity by defining ${r_1(x,y) \land r_2(y,z) \Rightarrow r_1(x,z)}$ which cannot be captured by models that do capture composition.
Nevertheless, this assumption does not lead to loss of generality for our study of generalized inference patterns, since we are allowed to use many rules, which in turn, can easily simulate cases that are excluded. For instance, the following rules:
$r_1(x,y) \land r_2(y,z) \Rightarrow r_3(x,z)$, and $r_3(x,y) \Rightarrow r_1(x,y)$ together simulate the transitivity rule given above.

We prove the statements in \Cref{thm:genPat} in two seperate parts. First, we prove the results given for BoxE from \Cref{tab:infPat}, and then we show the results given for the other models from \Cref{tab:infPat}.

\subsection{Proof of \Cref{thm:genPat}: BoxE}
\label{app:genInfProof}

We show that each generalized inference pattern can be captured by BoxE except for the composition pattern. For the latter, we argue why BoxE cannot capture this explicitly, as an inference pattern.

\paragraph{Generalized intersection.} We first introduce the concept of \emph{boxicity}. Let $G=(V,E)$ be a graph, where $V$ is the set of nodes, and $E$ is the set of edges. The \emph{boxicity} of $G$ is the minimum embedding dimension in which $G$ can be represented as an intersection of axis-aligned boxes, such that 
(i)~every box corresponds to a specific node, 
(ii)~boxes intersect iff an edge connects their respective nodes \cite{Roberts68}. 
It has been shown that the boxicity of a graph with $p$ edges is $O(\sqrt{p \cdot \log(p)})$ \cite{Chandran08}. This implies that, given a graph $G$, where every relation $r \in \Rbf$ is represented as a node in the graph, and every edge between them represents an intersection,
any finite combination of intersections between relations can be represented in a finite-dimensional vector space of worst-case dimensionality $O(|\Rbf| \sqrt{\log(|\Rbf|)})$. 

For a given knowledge graph, we define a \emph{relation intersection graph}. That is, for every relation $r \in \Rbf$, we define two nodes,  corresponding to its head and tail boxes, and then set edges in the graph based on desired intersections between relation boxes, which are dictated by intersection rules. 

Prior to encoding rules into relation intersection graph edges, we first compute the \emph{deductive closure} of the set of intersection rules. In other words, we check whether any rule of the form $r_i(x,y) \Rightarrow r_j(x,y)$, or $r_i(x,y) \land r_j(x,y) \Rightarrow r_k(x,y)$ for any $i,j,k$ can be entailed from the given set of rules, and keep adding new rules to this initial set, until no more rules can be deduced. That is, we compute the logical closure of the initial set. This allows us to make all possible intersections between relations explicit.

Then, we map all rules in the computed deductive closure to edges as follows: 
\begin{itemize}
\item For every intersection rule $r_1(x,y) \land r_2(x,y) \Rightarrow r_3(x,y)$, we set edges between the node corresponding to the head of $r_3$ and those of $r_1$ and $r_2$, with the same done for tail nodes. 
\item For every deduced hierarchy rule $r_1(x,y) \Rightarrow r_2(x,y)$, we set edges between the head nodes of $r_2$ and $r_1$, with the same done for tail nodes.
\end{itemize}

With the resulting relation intersection graph $G$, we have encoded necessary conditions for all rules to hold, namely that relations whose intersections are contained in other relations intersect with these relations. We now leverage the boxicity argument, and show that there exists a box configuration of finite dimensionality capturing all the intersections encoded in $G$. This box configuration captures all intersections needed between the respective boxes for the rules to hold, but is not necessarily sufficient to capture hierarchies and box containment. Hence, we modify the aforementioned box configuration using a procedure, which we apply iteratively over every intersection rule, such that the final configuration provably captures all rules, without capturing additional undesired rules. 

Our box reconfiguration procedure is as follows:
\begin{enumerate}
    \item Iterate over every intersection rule $r_1(x,y) \land r_2(x,y) \Rightarrow r_3(x,y)$:
    \begin{enumerate}[(a)]
    \item If the $r_3$ head and tail boxes do not contain the head and tail box intersections $r_1 \cap r_2$, then we grow these $r_3$ boxes by the minimum possible amount to make this condition hold and establish the rule. In other words, we grow the $r_3$ boxes at every position to equal the boundary of either the $r_1$ boxes or the $r_2$ boxes at the dimensions where the rule does not hold due to $r_1$ or $r_2$. This growth operation preserves all existing edges in $G$, and does not force new intersections, as all forced intersections due to rule capturing are already encoded by the existing edges.
    
    \item Following Part (a), the growth of $r_3$ can violate another rule in the set, in particular if $r_3$ is in the body of this rule. Hence, when any $r_3$ boxes are grown in Part (a), check all other intersection rules in the rule set: If the change in $r_3$ makes a rule no longer hold (i.e., the rule was captured prior to growing $\bm{r_3}$ and no longer is), then recursively call this procedure for this rule.
    \end{enumerate}

\end{enumerate}

We now show that this procedure is correct, and then prove that it terminates, particularly with respect to the number of recursive calls made. First, we note that, following a successful iteration on a given rule, a rule is successfully captured (Part (a)), and no other rules are violated in the process (Part (b)). Thus, the final configuration returned by this procedure over the initial boxicity-given configuration returns a valid BoxE configuration. In particular, this configuration captures all and only the provided patterns within the deductive closure, which includes the original intersection rules. Furthermore, growing $r_3$ boxes in the configuration to satisfy an intersection rule does not induce any rules outside their deductive closure. Indeed, when $r_3$ boxes are grown, they are only grown in dimensions where they fail to capture $\bm{r_1} \cap \bm{r_2}$. Thus, the procedure can only make $\bm{r_3}$ intersect with boxes that intersect with $\bm{r_1}$ or $\bm{r_2}$. As a result, the procedure can only force intersections between boxes within the deductive closure of the rule set. 

It now remains to show that this procedure terminates, and thus that a configuration of this kind indeed can be found. In particular, we study the maximal number of recursive calls needed. Consider a rule $\rho: r_1(x,y) \land r_2(x,y) \Rightarrow r_3(x,y)$, where $r_3$ boxes are grown. For simplicity, we only consider a single box  for $r_3$, i.e., a unique arity position, as the analysis is analogous at every arity position. We define \emph{boundaries} as being the lower and upper limits of a box at every dimension. Thus, a $d$-dimensional box has $2d$ boundaries. Therefore, in our binary BoxE configuration with $|\Rbf|$ relations and $d=O(|\Rbf|\sqrt{\log{|\Rbf|}})$, there are $O(|\Rbf|^2\sqrt{\log{|\Rbf|}})$ boundaries. For our analysis, we are interested in the number of \emph{distinct} boundaries in our configuration. 

We now consider the effect of an application of a call to Part (a) of the procedure on the number of distinct boundaries. If $\rho$ is already captured, then no action is needed. Otherwise, $\bm{r_3}$ needs to be grown. Hence, in this scenario, there exists at least one dimension in which the lower (resp., upper) boundary of $\bm{r_3}$ is strictly higher (resp., lower) than the maximum (resp., minimum) lower (resp., upper) bound of either $\bm{r_1}$ or $\bm{r_2}$. Therefore, when $\bm{r_3}$ is grown, the value of the problematic bound(s) at this dimension is made equal to the corresponding bound(s) of $\bm{r_1}$ or $\bm{r_2}$. As a result, the number of distinct boundaries is guaranteed to strictly drop by at least 1 following any growth operation. 

Furthermore, we consider the recursive calls made in Part (b), after any growth to $\bm{r_3}$. Observe that recursion is only called when the change to $\bm{r_3}$ exclusively makes the checked rule false. This condition ensures that all recursive calls are made only when the growing of the rule head boxes, in this case $\bm{r_3}$, is the exclusive cause for rule violation, and so eliminates all other possible causes of violation such that they are handled only when the outer loop iterating reaches the corresponding rule, and thus greatly simplifies the analysis. Finally, we observe that box growth can only be triggered when distinct boundaries exist. Hence, when the number of distinct boundaries drops to its (highly pessimistic and loose) minimum possible value of 1, no more recursive calls can be made. This observation, combined with the earlier finding that every box growth strictly reduces the number of distinct boundaries by at least 1, implies that the number of recursive calls in this procedure is upper bounded by $O(|\Rbf|^2\sqrt{\log{|\Rbf|}})$. Hence, this procedure terminates, and a BoxE configuration capturing generalized intersections exists.

\paragraph{Generalized hierarchy.} The proof for generalized intersection immediately applies to generalized hierarchies. 

\paragraph{Generalized symmetry.} The symmetry inference pattern is a single-relation pattern, and can appear at most once per relation. Symmetry can be easily captured for a relation $r$ by setting $\bm{r^{(1)}}$ and $\bm{r^{(2)}}$ to be identical boxes. This can be independently done for any relation, and thus BoxE captures generalized symmetry. 

\paragraph{Generalized anti-symmetry.} Analogously to generalized symmetry, anti-symmetry is a single-relation pattern. This pattern is captured by setting $\bm{r^{(1)}}$ and $\bm{r^{(2)}}$ to be disjoint for every anti-symmetric $r$. Therefore, BoxE captures generalized anti-symmetry.

\paragraph{Generalized inversion.} An inversion pattern $r_1(x,y) \Leftrightarrow r_2(y,x)$ can be captured by setting $\bm{r_1^{(1)}}$ and $\bm{r_2^{(2)}}$, as well as $\bm{r_1^{(2)}}$ and $\bm{r_2^{(1)}}$, to be identical boxes. This box sharing between inverse relations can easily be extended to any arbitrary set of inversion rules. 

\paragraph{Generalized mutual exclusion.} It is sufficient to observe that there exists a BoxE configuration for any arbitrary set of mutual exclusion rules due to the boxicity argument: simply consider a graph $G$ with no edges connecting mutually exclusive relations. A simpler argument can be given directly: generalized mutual exclusion can be achieved by making one of the relation boxes (head, or tail) disjoint in a fixed-dimensional space. 

\paragraph{(Generalized) composition.} Consider the composition pattern $r_1(x,y) \land r_2(y,z) \rightarrow r_3(x,z)$. In this pattern, we see that the entity that will appear in lieu of variable $x$ will be bumped differently in every atom, as it appears with different entities. More concretely, if we replace variables $x,y,z$ with entities $e_1, e_2, e_3$ respectively, then $\bm{e_1^{r_1}} = \bm{e_1} + \bm{b_2}$ and $\bm{e_1^{r_3}} = \bm{e_1} + \bm{b_3}$. We can also view bumps as equivalently applying to boxes, i.e., instead of $\bm{e_1} + \bm{b_2} \in \bm{r_1^{(1)}}$, we write $\bm{e_1} \in \bm{r_1^{(1)}} - \bm{b_2}$. Hence, it is equivalent to view BoxE as bumping relation boxes in the opposite direction. 

Now, we can see that  $\bm{r_1^{(1)}}$ is bumped by $-\bm{b_2}$, whereas $\bm{r_3^{(1)}}$ is bumped by $-\bm{b_3}$. Therefore, since bumps are entity-specific and unknown a priori since the bump stems from an abstract variable, one cannot analyze the relative positions of $\bm{r_1^{(1)}}$ and $\bm{r_3^{(1)}}$ and draw conclusions. By contrast, all other captured rules in BoxE are such that relation boxes corresponding to the same variable are bumped identically, which in effect neutralizes the effect of bumping and enables the capturing of the patterns. Hence, translational bumps, which allow BoxE to be fully expressive, prevent BoxE from capturing compositions.

\subsection{Proof of \Cref{thm:genPat}: Other models}

In what follows, we generally define KGC embedding models such that every KG entity is represented by a vector in $\Rbb^d$, and every relation defines two map functions $r_{h},r_{t}: \Rbb^d \rightarrow \Rbb^d$, which apply to head and tail embeddings, respectively. %
We further define the relation scoring function over a KG triple $s_r: \Rbb^d \times \Rbb^d \rightarrow \Rbb$ as a map from entity pair representations following the application of $r_h$ and $r_t$ to a real-valued score. %

\subsubsection{Translational models: TransE and RotatE}
We note that some of the results stated below are taken from the literature, but we included them nevertheless for completeness. The novel results are given for the generalized inference patterns.

For translational models, $r_h(\bm{e_1})$ encodes the translation (resp. rotation) operation, ${r_t(\bm{e_2}) = \bm{e_2}}$, and ${s_r(e_1, e_2) = \left\Vert r_t(\bm{e_2}) - r_h(\bm{e_1})\right\Vert}$.

\paragraph{Hierarchy.} Let $\Mmc_r = s_r^{-1}([0,\epsilon])$, where $s_r^{-1}$ is the inverse map of $s_r$, be the subset of embedding pairs $(v,w) \in \Rbb^d \times \Rbb^d$ such that $s_r(v, w) \leq \epsilon$, i.e., the decision region of the relation $r$ with margin $\epsilon$. 
As a result, $r_1(x,y) \Rightarrow r_2(x,y)$ holds iff  $\mathcal \Mmc_1 \subset \Mmc_2$. In TransE (resp., RotatE),  $(\bm {e_1}, \bm{e_2}) \in \mathcal \Mmc_r$  if $\bm {e_1}+\bm r - \bm {e_2} \in D_{\epsilon}(0)$ (resp., $\|  \bm {e_1} \circ \bm{r} -  \bm {e_2} \| \in D_{\epsilon}(0)$), where $D_\epsilon(0)$ is the disk of center $0$ and radius $\epsilon$. 
Since it is necessary that $M_1 \subset M_2$,  we require that the disk $D_{1,\epsilon}(\bm {e_1}+\bm {r_1})$ (resp., $D_{1,\epsilon}(\bm {e_1}\circ \bm {r_1})$) and radius $\epsilon$ is contained in the corresponding disk $D_2$, defined analogously using $r_2$. Since $D_1$ and $D_2$ have the same margin-induced radius, this is only possible if $r_1 = r_2$, effectively enforcing relation equivalence. Thus, neither translational model can capture hierarchies.

\paragraph{Intersection.} A model can represent the intersection pattern $r_1(x,y) \land r_2(x,y) \Rightarrow r_3(x,y)$ if $\mathcal M_{1} \cap \mathcal M_{2} \subset \mathcal M_{3}$. In TransE and RotatE, this is satisfied if $r_3$ lies in the centre of the disk intersection of $D_{\epsilon}(r_1)$ and $D_{\epsilon}(r_2)$, thus both models capture intersection. However, both models clearly fail to capture generalized intersection. In particular, if we consider rules ${r_1(x,y) \land r_2(x,y) \Rightarrow r_3(x,y)}$ and ${r_3(x,y) \land r_2(x,y) \Rightarrow r_1(x,y)}$, the rule ${r_2(x,y) \Rightarrow r_1(x,y)}$ is logically implied. But this is a hierarchy rule that clearly cannot be captured by either model. Hence, TransE and RotatE cannot capture generalized intersections.

\paragraph{Symmetry.} In TransE, $r(x,y) \Rightarrow r(y,x)$ holds iff $\bm{r} = 0$, which implies that $r$ is reflexive. Thus, TransE does not capture symmetry. 
In contrast, in RotatE, symmetry is captured iff ${r = \{\pm k\pi\}^d}$, ${k \in \Nbb}$, i.e., a rotation vector consisting exclusively of multiples of $\pi$. Symmetry is a single-relation pattern, and thus multiple rules, affecting different relations, can be captured independently. Hence, RotatE captures generalized symmetry. 

\paragraph{Anti-symmetry.} In TransE, a relation $r$ is anti-symmetric iff ${\| \bm r \| \geq \epsilon}$. The result for RotatE is proven in the original work \cite{RotatE-ICLR19}. As anti-symmetry is a single-relation pattern, it can be applied independently across all relations. Thus, both TransE and RotatE capture generalized anti-symmetry. 

\paragraph{Inversion.} For both TransE and RotatE, inversion holds iff $\bm{r_1} = - \bm{r_2}$. However, whereas RotatE can capture generalized inversion through repeated application of the earlier equation across all inversion rules, since it can handle any deduced symmetry results, TransE cannot. 
More concretely, consider the rule set ${r_1(x,y) \Leftrightarrow r_2(y,x)}$, ${r_2(x,y) \Leftrightarrow r_3(y,x)}$, ${r_3(x,y) \Leftrightarrow r_1(y,x)}$. This rule set implies ${r_1(x,y)  \Leftrightarrow r_1(y,x)}$, which RotatE can capture, but which TransE cannot. More generally, generalized inversion rules can yield symmetry rules, and thus only RotatE can capture generalized inversion.

\paragraph{Mutual exclusion.}
To capture mutual exclusion between relations $r_1$ and $r_2$, the model must satisfy ${\mathcal M_1 \cap \mathcal M_2 = \varnothing}$. In TransE, this holds iff  $\|\bm {r_1} - \bm {r_2}\| \geq 2\epsilon$. Analogously, for RotatE, this holds if $|\bm {r_i} - \bm {r_j}| \geq \arcsin(2\epsilon)$ at every dimension and all node embeddings have a norm of at least 1. Such constructions can be set up for arbitrarily many mutual exclusion pairs, through decreasing $\epsilon$ or increasing the magnitude of embeddings. Thus, both TransE and RotatE can capture generalized mutual exclusions.

\paragraph{Composition.} For TransE (resp., RotatE), two relations $r_1$ and $r_2$ compose a third relation $r_3$ iff  $\bm {r_1} + \bm {r_2} = \bm {r_3}$ (resp., $\bm{r_1} \circ \bm{r_2} = \bm{r_3})$. On the other hand, both fail to capture generalized compositions. In particular, for the rules $r_1(x,y) \land r_2(y,z) \Rightarrow r_3(x,z)$ and $r_1(x,y) \land r_4(y,z) \Rightarrow r_3(x,z)$, both models  force $r_2 = r_4$ (In RotatE, the equality is modulo $2\pi$). 

\subsubsection{Bilinear models: DistMult, ComplEx, TuckER}

TuckER is shown to subsume DistMult and ComplEx \cite{TuckER}, so all positive results for either ComplEx and DistMult automatically follow for TuckER.  Hence, these positive results for TuckER are omitted from the presentation. Analogously, when negative results are shown for TuckER, they automatically propagate to DistMult and ComplEx. 

We now formally introduce TuckER. TuckER learns a tensor ${\Wmc \in \Rbb^{d_e\times d_r \times d_e}}$,  $d_e,d_r \in \Nbb$, a vector $\bm  e \in \Rbb^{d_e}$ for every entity, and a vector ${\bm r \in \Rbb^{d_r}}$ for every relation, and ${s_r(\bm {e_1}, \bm{e_2}) = \mathcal W \cdot \bm  e_1 \cdot \bm r \cdot\bm  e_2}$. For ease of notation, we define ${v_{1,r} = \mathcal W \times \bm{e_1} \times \bm r}$. The scoring function can then be written as $s_r(\bm {e_1}, \bm{e_2}) = v_{r,1} \cdot \bm  e_2$. Given a head entity $e_1$ and a relation $r$, we define the space ${A_{1,r} = \{ \bm x  \in \Rbb^{d_e} \ | \ v_{r,1} \cdot \bm x \geq \epsilon \}}$.

\paragraph{Hierarchy.} For bilinear models, it has been shown that individual hierarchies can be captured, but not generalized hierarchies \cite{Gutirrez18}. In particular, to satisfy the rules ${r_1(x,y) \Rightarrow r_3(x,y)}$ and ${r_2(x,y) \Rightarrow r_3(x,y)}$ simultanously, bilinear models must set either ${r_1(x,y) \Rightarrow r_2(x,y)}$ or ${r_2(x,y) \Rightarrow r_1(x,y)}$.

\paragraph{Intersection.} We show that TuckER cannot capture intersections. In TuckER, a rule of the form ${r_1(x, y) \land r_2(x, y) \Rightarrow r_3(x, y)}$ holds iff ${A_{r_2,1} \cap A_{r_1,1} \subset A_{r_3,1}}$, ${\forall \bm e \in \Rbb^{d_e}}$. This is true iff $v_{r_1,1}, v_{r_2,1}, v_{r_3,1}$ are colinear, and thus that $r_1, r_2$, and $r_3$ are colinear. However, this also implies that either $r_1(x,y) \Rightarrow r_2(x,y)$, or $r_2(x,y) \Rightarrow r_1(x,y)$. Hence, TuckER fails to capture intersections. 

\paragraph{Symmetry.} ComplEx captures symmetry patterns by having real-only embedding matrices for its relations. DistMult is inherently symmetric by construction. Since symmetry is a single-relation pattern, multiple symmetries can be independently captured, and thus all three models can capture generalized symmetry.

\paragraph{Anti-symmetry.} DistMult cannot capture anti-symmetry, as it is inherently a symmetric model. ComplEx captures anti-symmetry by having imaginary-only embedding matrices for its relations. Analogously to symmetry, anti-symmetry is also a single-relation pattern, and thus ComplEx (and TuckER) can capture generalized anti-symmetry.

\paragraph{Inversion.} It is known that DistMult cannot capture inversions, while ComplEx can \cite{RotatE-ICLR19}. %
Generalized inversion can also be captured in ComplEx, as symmetry, the only other type of rule deducible from multiple inversions, is also captured by ComplEx. 

\paragraph{Mutual exclusion.} In TuckER, two relations $r_1$ and $r_2$ are mutually exclusive iff $r_1 = -r_2$. %
This implies TuckER can capture mutual exclusion, but cannot capture generalized mutual exclusions. In particular, to satisfy $r_1(x,y) \land r_2(x,y) \Rightarrow \bot$ and $r_1(x,y) \land r_3(x,y) \Rightarrow \bot$, TuckER forces $r_2 = r_3$.

\paragraph{Composition.} It is shown that both ComplEx and DistMult cannot capture composition patterns \cite{RotatE-ICLR19, Gutirrez18}. Furthermore, it is also known that relation maps must be bijective to be able to represent composition \cite{RotatE-ICLR19}. 
This is not the case in TuckER, as relations are surjective maps from $\Rbb^{d_e \times d_r}$ to $\Rbb^{d_e}$, and linear bijections between vector spaces are only possible with the same dimensionality. Hence, TuckER also cannot capture compositions.

\section{Proof of \Cref{thm:InfPat} (Inference Patterns as Rule Languages)}
\label{app:lang}

We show that a BoxE model of dimensionality $d = O(|\Rbf|^2)$ captures the rule language specified in \Cref{thm:InfPat}. This is achieved by leveraging the ideas from the generalized inference patterns proof in \Cref{app:infPat}. Indeed, our existence proof also builds on the boxicity argument used in this proof.

Let $S$ be a set of rules, and let $S_{p}$, $S_a$, and $S_m$ be subsets of $S$, where $S_p$ consists of hierarchy, symmetry, inversion, and intersection rules, $S_a$ consists of  anti-symmetry rules, and $S_m$ consists of mutual exclusion rules. We first show that rules from $S_p \cup S_a$ can be captured, then extend this to additionally capture $S_m $.

\paragraph{Step 1: Defining the relation intersection graph.} We define a set of $2|\Rbf|$ nodes, where every relation is encoded with 2 nodes for its head and tail boxes. We now constrain this graph to eventually capture all rules in $S_p$. First, we capture all symmetry and inversion rules as follows:
\begin{enumerate}
    \item \textbf{Symmetry:} For every symmetry rule, we combine the corresponding head and tail nodes of a relation $r$ to a single node. In other words, a single relation $r$  is made symmetric by encoding both $\bm{r^{(1)}}$ and $\bm{r^{(2)}}$ with one same node. This encoding enforces that the head and tail boxes of $r$ are identical, and thus that $r$ is indeed symmetric, as required. 
    \item \textbf{Inversion:} For every inversion rule $r_1(x,y) \Rightarrow r_2(y,x)$, we combine the respective head and tail nodes of $r_1$ and $r_2$ such that $\bm{r_{1}^{(1)}}$ and $\bm{r_{2}^{(2)}}$, as well as $\bm{r_{1}^{(2)}}$ and $\bm{r_{2}^{(1)}}$, are each represented by one node. This makes that their corresponding boxes are equal, effectively capturing inversion patterns. 
\end{enumerate}

Following this step, $G$ now consists of at most $2|\Rbf|$ nodes, and captures symmetry and inversion rules jointly. It now remains to define edges in $G$, as needed to later capture intersection and hierarchy rules. This is done analogously to the proof for generalized intersections (cf. \Cref{app:genInfProof}): First, the deductive closure of all intersection and hierarchy rules is computed, and the corresponding edges are encoded in $G$. Note that the resulting graph $G$ continues to capture inversion and symmetry, as these rules are encoded through nodes, and also encodes the deductive closure of all rules in $S_p$. Indeed, any box intersection imposed by the deductive closure of intersection and hierarchy rules with a node capturing a symmetry or inversion rule automatically implies a box intersection with the multiple boxes that the node represents. Hence, $G$ enables capturing symmetry and inversion rules a priori, as well as jointly sets up the necessary edges for hierarchy and inversion rules. Finally, we leverage the boxicity argument, and our final graph $G$, to obtain a box configuration where all the box intersections needed to later capture hierarchy and intersection rules are present (but not necessarily capturing hierarchy and intersection patterns at this stage), and which also successfully captures inversion and symmetry rules.

\paragraph{Step 2: Anti-symmetry ($S_a$).} Anti-symmetry rules are captured by adding additional dimensions to the box configuration resulting from Step 1 to distinguish between the head and tail boxes of an anti-symmetric relation. $S$ is consistent, therefore only anti-symmetry rules not contradicting the set of rules $S_p \cup S_m$ can be given. For example, if symmetry rule $r(x,y) \Leftrightarrow r(y,x) \in S_{p}$, then $r(x,y) \Rightarrow \neg r(y,x) \notin S_a$. This is important, as it implies that no combination of hierarchy, inversion, intersection symmetry, and mutual exclusion rule can force an intersection between $\bm{r^{(1)}}$ and $\bm{r^{(2)}}$, for any anti-symmetric $r$, and thus, that subsequent steps in this proof preserve the anti-symmetry captured in this step.

We now capture anti-symmetry rules by dedicating a new ``disjointness'' dimension for all boxes, such that, for an anti-symmetric relation $r$, the box ranges for head and tail boxes are made disjoint in this dimension, i.e., $[l^{(1)},u^{(1)}] \cap [l^{(2)},u^{(2)}] = \phi$, and are set arbitrarily for all other relations, such that, for all rules in $S_p$, if an anti-symmetric $r$ is the head of a hierarchy rule $r_1 \implies r$, then the ranges of $\bm{r_1}$ in this dimension respect the hierarchy and, for an intersection rule $r_1 \land r_2 \implies r$, then $\bm{r_1} \cap \bm{r_2} \subset \bm{r}$. This initialization exists, as $S$ is consistent, so cannot create conflicting interval requirements for relations in rules. One can also observe this by considering this initialization a recursive pass through the rule sets affected by the anti-symmetric relations, where all other uninitialized relations in the deductive closure are not yet set. Hence, this new dimension captures $\bm{r^{(1)}} \cap \bm{r^{(2)}} = \phi$, so correctly captures anti-symmetry, and cannot be broken by subsequent rule-based box growth. It also is compatible with all symmetry and inversion rules, as box sharing is maintained. %
Hence, our current BoxE configuration captures any consistent set of anti-symmetry, symmetry, and inversion rules. Given $S_a$, at most $|S_a|$ additional dimensions are needed, and since at most $|\Rbf|$ anti-symmetry rules can exist, the worst-case dimensionality of our configuration remains ${O(|\Rbf|\sqrt{\log{(|\Rbf|)}})}$.
We now build on this result and show that the current configuration can be modified to additionally capture intersection and hierarchy rules.

\paragraph{Step 3: Hierarchies and intersections.} Given the box configuration at the end of Step 2, we now apply the box reconfiguration procedure presented in the generalized intersections proof (cf. \Cref{app:genInfProof}) to capture all hierarchy and intersection rules in $S$. We also note that, since $S$ is consistent, no hierarchy and intersection rules force any inconsistency with the already captured symmetry, anti-symmetry and inversion rules, e.g., if $r_1(x,y) \Rightarrow r_1(y,x), r_2(x,y) \Rightarrow \neg r_2(y,x) \in S$, then $r_1(x,y) \Rightarrow r_2(x,y) \notin S$. Thus all symmetry, anti-symmetry, and inversion patterns, whose capture is based on structural concepts (box sharing and dedicated dimensions respectively), are preserved. In particular, box sharing is unaffected, and no box growth from this step can break the disjointness of anti-symmetric relation boxes, as $S$ is consistent. The completeness of the procedure with respect to hierarchy and intersection rules is also shown in \Cref{app:genInfProof}.

\paragraph{Step 4: Mutual exclusion.} %

Given the BoxE configuration from Step 3, capturing rules from $S_{p} \cup S_{a}$, we also capture rules from $S_m$ with additional dimensions. Indeed, we show that this can be done using a BoxE configuration with $d = O(|\Rbf|^2)$ dimensions. Starting from the configuration after the completion of Step 3, we now dedicate a single dimension per mutual exclusion rule, and capture this pattern as follows: For every mutual exclusion rule, we set a dimension, where $r_1$ and $r_2$ have disjoint range intervals $z_1, z_2 \subset [0,1]$, such that, without loss of generality, $z_1 = [z_{1,\text{min}}, z_{1,\text{max}}]$, $z_2 = [z_{2,\text{min}}, z_{2,\text{max}}]$ and $z_{2,\text{min}} > z_{1,\text{max}}$.  Then, we set the range of every other box in the configuration at this new dimension analogously to Step 2 (i.e., arbitrarily, but in a rule-aware fashion) by repeating the box reconfiguration procedure in Step 3 for capturing hierarchy and intersection rules starting from the current configuration. 

Note that anti-symmetry, symmetry, and inversion rules play no part in this step, as anti-symmetry rules are captured with dedicated dimensions as shown earlier, whereas symmetry and inversion rules are already enforced, and thus captured, through box sharing and equality.

Intuitively, this step first makes $r_1$ and $r_2$ mutually exclusive in one dimension, then recursively traverses the set of hierarchy and intersection rules, as in Step 3, to preserve the capturing of these rules in this new dimension specifically. Clearly, anti-symmetry remains true, since its dedicated dimension is not affected by the repetition of Step 3. Furthermore, since $S$ is consistent, all mutual exclusion rules in $S$ can be captured without causing inconsistency. In other words, rule sets such as $r_1(x,y) \Rightarrow r_2(x,y), r_1(x,y) \Rightarrow r_3(x,y), $ and $r_2(x,y) \land  r_3(x,y) \Rightarrow \bot$ are not possible. 

Hence, since $|S_m| \leq 0.5|\Rbf|(|\Rbf| -1)$, the number of distinct pairs that can be selected from \Rbf, a BoxE model with $d = 0.5|\Rbf|(|\Rbf| -1) + |\Rbf| + |\Rbf|\sqrt{\log{|\Rbf|}} = O(|\Rbf|^2)$ dimensions can capture any consistent set of rules $S$ from the language of intersection, hierarchy, symmetry, anti-symmetry, mutual exclusion, and inversion rules.

We finally highlight one subtle, but important detail: Whereas the inference pattern language just described can be captured by a BoxE model having $d=O(|\Rbf|^2)$ dimensions, some individual generalized patterns (inversion, hierarchy, symmetry, anti-symmetry, mutual exclusion) can be captured with even constant number of dimensions, and generalized intersection can be captured with $O(|\Rbf|\sqrt{\log{(|\Rbf|)}})$ dimensions. %
Hence, an interesting contrast in dimensionality requirements arises between capturing individual generalized inference patterns, capturing the language of \Cref{thm:ruleInj}, and capturing rule language of \Cref{thm:InfPat}, which highlights the significantly larger requirements that capturing joint generalized requirements, and the potential existence of cycles, can impose on any embedding model.

\section{Proof of Theorem \ref{thm:ruleInj} (Rule Injection)}
\label{app:ruleInj}

We now prove that arbitrary sets $S_p$ of hierarchy, intersection, symmetry, and inversion rules can be injected into BoxE. To this end, we adapt the proof of \Cref{thm:InfPat} to this setting.%

We start with a randomly initialized box configuration. First, we inject inversion and symmetry rules %
using box sharing: For symmetry rules, we set $\bm{r^{(1)}} = \bm{r^{(2)}}$, and for inversion rules, we set $\bm{r_1^{(1)}}
= \bm{r_2^{(2)}}$ and $\bm{r_2^{(1)}} = \bm{r_1^{(2)}}$, and this can be done in linear time with respect to the number of inversion and symmetry rules. This achieves the same result as the node sharing in Step 1 of the proof of \Cref{thm:InfPat}, except that the box configuration is a concrete random initialization, as opposed to an abstract configuration known to exist due to boxicity. We then proceed with the box reconfiguration procedure in Step 3 of this same proof to enforce hierarchy and intersection rules on top of inversion and symmetry rules. This step is guaranteed to enforce these rules, and their deductive closure, as shown in \Cref{app:lang}, and maintains box sharing, so preserves symmetry and inversion. 

We now analyze the worst-case runtime complexity of the box reconfiguration procedure. We assume the worst-case, that any pairwise intersections should be expressible, and thus use a dimensionality $d = O(|\Rbf|\sqrt{\log{(|\Rbf|}})$. The worst-case running time of the box reconfiguration procedure 
for enforcing a single hierarchy/intersection rule is $O(|\Rbf|d) = O(|\Rbf|^2\sqrt{\log{(|\Rbf|}})$, corresponding to the maximum number of boundary changes needed per call. However, this upper bound is independent of the number of rules in $S$, as no more than $O(|\Rbf|d)$ steps can be made across all rules. Thus, the worst-case running time for rule injection across all hierarchy and intersection rules is $O(|\Rbf|d) = O(|\Rbf|^2\sqrt{\log{(|\Rbf|)}})$. 

Hence, rule injection for hierarchy and intersection rules runs at worst in near-quadratic time with respect to %
$|\Rbf|$, a typically small number, irrespective of the number of these rules. This result, combined with the efficiency of enforcing symmetry and hierarchy, imply that BoxE can be efficiently injected with arbitrary sets of symmetry, inversion, hierarchy and intersection rules.

\section{Experimental Details}
In this section, we give further details on the experiments that we have conducted. In particular, we report details of every dataset, the hyperparameter tuning setup used when training BoxE, as well as the final set of hyperparameters used in the configurations whose results we report in the paper. Finally, we report the complete set of results for KGC, higher-arity, and rule injection experiments, i.e., MR, MRR, Hits@1, Hits@3, and Hits@10. All reported results for the KGC and KBC experiments are average results from 3 training runs, and empirically have very small variance. In particular, all MRR values fluctuate by no more than $0.002$ between runs across all datasets. 

\label{app:exp}
\subsection{Benchmark dataset details}
\begin{table}[t!]
	\centering
	\caption{Properties of benchmark datasets FB15k-237, WN18RR, YAGO3-10, JF17K, and FB-AUTO.} 
	\begin{tabular}{lccccc}
		\toprule 
		 {Dataset} & $|\Ebf|$ & $|\Rbf|$ & Training Facts & Validation Facts & Testing Facts\\
		\cmidrule(r){2-6}
		 FB15k-237 & 14,541 & 237 & 272,115 & 17,535 & 20,466\\
		 WN18RR & 40,943 & 11 & 86,835 & 3,034 & 3,034 \\
		 YAGO3-10 & 123,182 & 37 & 1,079,040 & 5,000 & 5,000 \\
		 JF17K & 29,257 & 327 & 61,911 & 15,822 & 24,915 \\
		 FB-AUTO & 3,388 & 8 & 6,778 & 2,255 & 2,180 \\
		\bottomrule
	\end{tabular}
	\label{tab:DatasetDetails}
\end{table}

In this subsection, we provide the details of of all benchmark datasets used in this paper (FB15k-237, WN18RR, YAGO3-10, JF17K, and FB-AUTO), namely the number of entities, relations, and facts in every split (training, validation, and test) in \Cref{tab:DatasetDetails}.  

\subsection{Hyperparameter settings for BoxE experiments}
BoxE is trained using the Adam optimizer \cite{Kingma-ICLR2014}, to optimize negative sampling loss \cite{RotatE-ICLR19}. Training for every run was conducted on a Haswell CPU node with 12 cores, 64 GB RAM, and a V100 GPU. Hyperparameter tuning was conducted over its learning rate $\lambda$, dimensionality $d$, loss margin $\gamma$, distance order $x$, and number of negative examples $m$. For all BoxE experiments, points and boxes were projected into the hypercube $[-1,1]^d$, a bounded space, by simply applying the hyperbolic tangent function $\tanh$ element-wise on all final embedding representations. 

Learning rate was varied between $10^{-6}$ and $10^{-2}$, with root values of 1,2,5 and exponents from -6 to -2, i.e., $10^{-6}, 2\times10^{-6}, 5\times10^{-6},$ etc. .
Margin was varied between 3 and 24 inclusive, in increments of 1.5, and in increments of 1 between 3 and 6. Adversarial temperature was varied between the integer values of 1 and 4 inclusive, and the number of negative samples was varied between 50, 100, and 150. 
Across all knowledge graph datasets, we additionally ran experiments with \emph{data augmentation}, such that, for every relation $r$, a distinct inverse relation $r'$ is defined, and every fact $r(e_1, e_2)$ is augmented with another fact $r'(e_2,e_1)$. This setting, however, was only marginally beneficial on YAGO3-10, yielding a slightly improved MR. 

Finally, the distance order was set to either 1 (Manhattan distance) or 2 (Euclidian distance), and batch sizes (for number of positive examples) were varied between all powers of two between $2^6$ and $2^{12}$ inclusive. Hyperparameters were initially selected randomly and tuned using grid search. The set of used hyperparameters in experiments is shown in \Cref{tab:HPSettings}. 

Aside from the reported hyperparameter settings, we have also attempted  to fix box sizes, either in a hard fashion or softly by setting maximum total size. Hard sizes were based on statistical popularity of relations, whereas soft totals were tuned. However, neither of these settings yielded any improvements, and in fact both have been mostly detrimental to performance. This, in fact, further highlights the importance of box size variability to obtaining good predictive performance. Interestingly, it also confirms that statistical popularity alone is not sufficient to establish optimal box sizing. We also remain very confident that BoxE performance can further improve in the future, as more dedicated empirical studies and more comprehensive and bespoke tuning methods are applied. 
\begin{table}[t!] 
	\centering
	\caption{Hyperparameter settings of BoxE over different datasets.} 
	\label{tab:HPSettings} 
	\small\addtolength{\tabcolsep}{-1pt}
	\begin{tabular}{l@{\hskip 6pt}c@{\hskip 4pt}c@{\hskip 4pt}c@{\hskip 4pt}c@{\hskip 4pt}c@{\hskip 4pt}c@{\hskip 4pt}c@{\hskip 4pt}c@{\hskip 4pt}}
		\toprule 
		Dataset & \makecell[l]{Embedding \\ Dimension} & Margin & \makecell[l]{Learning \\ Rate} & \makecell[l]{Adversarial \\ Temperature} & \makecell[l]{Negative \\Samples } & \makecell[l]{Distance \\ Order} & \makecell[l]{Batch \\ Size} & \makecell[l]{Data \\ Augmentation}\\
		\cmidrule{2-9}
		FB15k-237(u) & 500 & 12 & $1\times10^{-4}$ & 0.0 & 100 & 1 & 1024 & No\\
		FB15k-237(a) & 1000 & 3 & $5\times10^{-5}$ & 4.0 & 100 & 2 & 1024 & No\\
		WN18RR(u) & 500 & 5 & $1\times10^{-3}$ & 0.0 & 150 & 2 & 512 & No\\
		WN18RR(a) & 500 & 3 & $1\times10^{-3}$ & 2.0 & 100 & 2 & 512 & No\\
		YAGO3-10(u) & 200 & 10.5 & $1\times10^{-3}$ & 0.0 & 150 & 2 & 4096 & Yes\\
		YAGO3-10(a) & 200 & 6 & $1\times10^{-3}$ & 2.0 & 150 & 2 & 4096 & Yes\\
		JF17K(u) & 200 & 15 & $2\times10^{-3}$ & 0.0 & 100 & 2 & 1024 & N/A\\
		JF17K(a) & 200 & 5 & $1\times10^{-4}$ & 2.0 & 100 & 2 & 1024& N/A\\
		FB-AUTO(u) & 200 & 18 & $2\times10^{-3}$ & 0.0 & 100 & 2 & 1024 & N/A\\
		FB-AUTO(a) & 200 & 9 & $5\times10^{-4}$ & 2.0 & 100 & 2 & 1024 & N/A\\
		SportsNELL & 200 & 6 & $1\times10^{-3}$ & 0.0 & 100 & 2 & 1024 & No\\
		SportsNELL+RI & 200 & 6 & $1\times10^{-3}$ & 0.0 & 100 & 2 & 1024& No\\
		\bottomrule
	\end{tabular}
\end{table}

\subsection{Complete experimental results}
The complete results for KGC experiments on FB15k-237, WN18RR, and YAGO3-10 are reported across Tables \ref{tab:CompleteKGC} and \ref{tab:CompleteKGCYAGO}. Complete results for higher-arity KBC experiments on JF17K and FB-AUTO are reported in \Cref{tab:CompleteKBC}, and complete rule injection results for BoxE and BoxE+RI on the two SportsNELL evaluation sets are reported in \Cref{tab:CompleteRuleInj}.
\begin{table}[t!] 
	\centering
	\caption{Complete KGC results for BoxE and competing models on FB15K-237 and WN18RR.} 
	\label{tab:CompleteKGC} 
	\begin{tabular}{l@{\hskip 5pt}c@{\hskip 3pt}c@{\hskip 3pt}c@{\hskip 3pt}c@{\hskip 3pt}c@{\hskip 6pt}c@{\hskip 3pt}c@{\hskip 3pt}c@{\hskip 3pt}c@{\hskip 3pt}c@{\hskip 3pt}HHHHH}
		\toprule 
				 {Model} & \multicolumn{5}{c}{\textbf{FB15K-237}} &  \multicolumn{5}{c}{WN18RR} & \multicolumn{5}{c}{} \\
		\cmidrule(r){2-6}
		\cmidrule(r){7-11}
		 & MR & MRR & H@1 & H@3 & H@10 & MR & MRR & H@1 & H@3 & H@10 & MR & MRR & H@1 & H@3 & H@10\\
		 TransE(u)  \cite{ruffinelli2020you} & - & .313 & - & - & .497 & - & .228 & - & - & .520 & \textit{-} & \textit{-} & \textit{-} & \textit{-} & \textit{-} \\
		 RotatE(u) \cite{RotatE-ICLR19} & 185 & .297 & .205 & .328 & .480 & \textit{3254} & \textbf{\textit{.470}} & \textbf{\textit{.422}} & \textbf{\textit{.488}} & \textbf{\textit{.564}} & \textit{\textbf{1116}} & \textit{.459} & \textit{.360} & \textit{.509} & \textit{.651}\\
		 BoxE(u) & \textbf{172} & \textbf{.318} & \textbf{.223} & \textbf{.351} & \textbf{.514} & \fbox{\textbf{3117}} & .442 & .398 & .461 & .523 & 1164 & \fbox{\textbf{.567}} & \fbox{\textbf{.494}} & \fbox{\textbf{.611}} & \fbox{\textbf{.699}}\\
		 \midrule 
		 TransE(a) \cite{RotatE-ICLR19} & 170 & .332 & .233 & .372 & .531 & 3390 & .223 & .013 & .401 & .529 & - & - & - & - & -\\
		 RotatE(a) \cite{RotatE-ICLR19} & 177 & \textbf{.338} & \textbf{.241} & \textbf{.375} & .533 & 3340 & \fbox{\textbf{.476}} & \textbf{.428} & \fbox{\textbf{.492}} & \fbox{\textbf{.571}} & 1767 & .495 & .402 & .550 & .670\\
		 BoxE(a) & \fbox{\textbf{163}} & .337 & .238 & .374 & \textbf{.538} & \textbf{3207} & .451 & .400 & .472 & .541 & \fbox{\textbf{1022}} &  \textbf{.560} & \textbf{.484} & \textbf{.608}  & \textbf{.691}\\
		 \midrule 
		 DistMult \cite{ruffinelli2020you,DistMult-ICLR15} & - & .343 & - & - & .531 & - & .452 & - & - & .531 & 5926 & .34 & .24 & .38 & .54\\
		 ComplEx \cite{ruffinelli2020you,DistMult-ICLR15} & - & .348 & - & - & .536 & - & \textbf{.475} & - & - & \textbf{.547} & 6351 & .36 & .26 & .40 & .55\\
		 TuckER \cite{TuckER} & - & \fbox{\textbf{.358}} & \fbox{\textbf{.266}} & \fbox{\textbf{.394}} & \fbox{\textbf{.544}} & - & .470 & \fbox{\textbf{.443}} & \textbf{.482} & .526 & \textbf{\textit{4423}} & \textbf{\textit{.529}} & \textbf{\textit{.451}} &  \textbf{\textit{.576}} & \textbf{\textit{.670}}\\
		\bottomrule
		\bottomrule
	\end{tabular}
\end{table}

\begin{table}[t!] 
	\centering
	\caption{Complete KGC results for BoxE and competing models on YAGO3-10.}
	\label{tab:CompleteKGCYAGO} 
	\begin{tabular}{l@{\hskip 5pt}HHHHHHHHHHc@{\hskip 3pt}c@{\hskip 3pt}c@{\hskip 3pt}c@{\hskip 3pt}c@{\hskip 3pt}}
		\toprule 
		 {Model} & \multicolumn{5}{c}{} & \multicolumn{5}{c}{} & \multicolumn{5}{c}{\textbf{YAGO3-10}} \\
		\cmidrule(r){12-16}
		 & MR & MRR & H@1 & H@3 & H@10 & MR & MRR & H@1 & H@3 & H@10 & MR & MRR & H@1 & H@3 & H@10\\
		 TransE(u)  \cite{ruffinelli2020you} & - & .313 & - & - & .497 & - & .228 & - & - & .520 & \textit{-} & \textit{-} & \textit{-} & \textit{-} & \textit{-} \\
		 RotatE(u) \cite{RotatE-ICLR19} & 185 & .297 & .205 & .328 & .480 & \textit{3254} & \textbf{\textit{.470}} & \textbf{\textit{.422}} & \textbf{\textit{.488}} & \textbf{\textit{.564}} & \textit{\textbf{1116}} & \textit{.459} & \textit{.360} & \textit{.509} & \textit{.651}\\
		 BoxE(u) & \textbf{172} & \textbf{.318} & \textbf{.223} & \textbf{.351} & \textbf{.514} & \fbox{\textbf{3117}} & .442 & .398 & .461 & .523 & 1164 & \fbox{\textbf{.567}} & \fbox{\textbf{.494}} & \fbox{\textbf{.611}} & \fbox{\textbf{.699}}\\
		 \midrule 
		 TransE(a) \cite{RotatE-ICLR19} & 170 & .332 & .233 & .372 & .531 & 3390 & .223 & .013 & .401 & .529 & - & - & - & - & -\\
		 RotatE(a) \cite{RotatE-ICLR19} & 177 & \textbf{.338} & \textbf{.241} & \textbf{.375} & .533 & 3340 & \fbox{\textbf{.476}} & \textbf{.428} & \fbox{\textbf{.492}} & \fbox{\textbf{.571}} & 1767 & .495 & .402 & .550 & .670\\
		 BoxE(a) & \fbox{\textbf{163}} & .337 & .238 & .374 & \textbf{.538} & \textbf{3207} & .451 & .400 & .472 & .541 & \fbox{\textbf{1022}} &  \textbf{.560} & \textbf{.484} & \textbf{.608}  & \textbf{.691}\\
		 \midrule 
		 DistMult \cite{ruffinelli2020you,DistMult-ICLR15} & - & .343 & - & - & .531 & - & .452 & - & - & .531 & 5926 & .34 & .24 & .38 & .54\\
		 ComplEx \cite{ruffinelli2020you,DistMult-ICLR15} & - & .348 & - & - & .536 & - & \textbf{.475} & - & - & \textbf{.547} & 6351 & .36 & .26 & .40 & .55\\
		 TuckER \cite{TuckER} & - & \fbox{\textbf{.358}} & \fbox{\textbf{.266}} & \fbox{\textbf{.394}} & \fbox{\textbf{.544}} & - & .470 & \fbox{\textbf{.443}} & \textbf{.482} & .526 & \textbf{\textit{4423}} & \textbf{\textit{.529}} & \textbf{\textit{.451}} &  \textbf{\textit{.576}} & \textbf{\textit{.670}}\\
		\bottomrule
	\end{tabular}
\end{table}

\begin{table}[h!] 
	\centering
	\caption{Complete KBC results %
	on higher-arity datasets JF17K and FB-AUTO.} 
	\label{tab:CompleteKBC} 
	\small\addtolength{\tabcolsep}{-1pt}
	\begin{tabular}{lcccccccccc}
		\toprule 
		\multirow{2}{*}{Model} & \multicolumn{5}{c}{\textbf{JF17K}} & \multicolumn{5}{c}{\textbf{FB-AUTO}} \\
		\cmidrule(r){2-6}
		\cmidrule(r){7-11}
		 & MR & MRR & H@1 & H@3 & H@10 & MR & MRR & H@1 & H@3 & H@10\\
		 m-TransH & - & .446 & .357 & .495 & .614 & - & .728 & .727 & .728 & .728\\
		 m-DistMult & - & .460 & .367& .510 & .635 & - & .784 & .745 & .815 & .845\\
		 m-CP & - & .392 & .303 & .441& .560 & - & .752 & .704 & .785 & .837\\
		 HypE & - & .492 & .409 & .533 & .650 & - & .804 & .774 & .823 & .856\\
		 HSimplE & - & .472 & .375 & .523 & .649 & - & .798 & .766 & .821 & .855\\
		 BoxE(u) & \fbox{\textbf{363}} & .553 & .467 & .596 & .711 & \fbox{\textbf{110}} & .837 & .804 & .858 & .895\\
		 BoxE(a) & 372 & \fbox{\textbf{.560}} & \fbox{\textbf{.472}} & \fbox{\textbf{.604}} & \fbox{\textbf{.722}} & 122 & \fbox{\textbf{.844}} & \fbox{\textbf{.814}} & \fbox{\textbf{.863}} & \fbox{\textbf{.898}}\\
		\bottomrule
	\end{tabular}
\end{table}

\begin{table}[t!]
	\centering
	\caption{Complete rule injection experiment results on the SportsNELL full and filtered sets.} %
	\label{tab:CompleteRuleInj} 
	\begin{tabular}{l@{\hskip 5pt}c@{\hskip 3pt}c@{\hskip 3pt}c@{\hskip 3pt}c@{\hskip 3pt}c@{\hskip 5pt}c@{\hskip 3pt}c@{\hskip 3pt}c@{\hskip 3pt}c@{\hskip 3pt}c@{\hskip 3pt}}
		\toprule 
		Model & \multicolumn{5}{c}{Full Set} & \multicolumn{5}{c}{Filtered  Set} \\
		\cmidrule(r){2-6}\cmidrule(r){7-11}
		 & MR & MRR & H@1 & H@3 & H@10 & MR & MRR & H@1 & H@3 & H@10\\
		 BoxE & 17.4 & .577 & .478 & .623 & .780 & 19.1 & .713 & .661 & .732 & .824\\
		 BoxE+RI & \fbox{\textbf{1.74}} & \fbox{\textbf{.979}} & \fbox{\textbf{.968}} & \fbox{\textbf{.988}} & \fbox{\textbf{.997}} & \fbox{\textbf{5.11}} & \fbox{\textbf{.954}} & \fbox{\textbf{.938}} & \fbox{\textbf{.964}} & \fbox{\textbf{.984}} \\
		\bottomrule
	\end{tabular}
\end{table}
\section{Additional Experimental Insights and Discussions}
\label{app:expInsight}

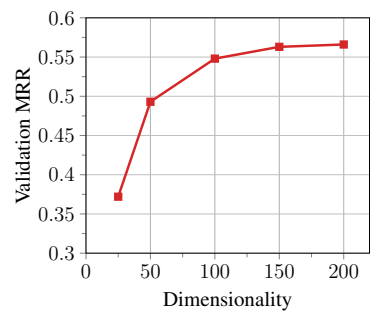
\begin{wrapfigure}[12]{r}{0.4\textwidth}
\centering
\vspace{-0.48cm}
\begin{tikzpicture}[scale=0.55]
\definecolor{color0}{rgb}{0.83921568627451,0.152941176470588,0.156862745098039}
\pgfplotsset{every tick label/.append style={font=\Large}}
\begin{axis}[
tick align=outside,
tick pos=left,
x grid style={white!69.01960784313725!black},
xlabel={\Large Dimensionality},
 x label style={at={(axis description cs:0.5,-0.05)},anchor=north},
y label style={at={(axis description cs:0,.5)},anchor=south},
xmajorgrids,
xmin=0, xmax=220,
ylabel={\Large Validation MRR},
ymajorgrids,
ymin=0.3, ymax=0.6,
minor tick num=1,
xtick={0,50,100,150,200},ytick={0.3,0.35,0.4,0.45,0.5,0.55,0.6},
]
\addplot [line width=0.6mm, color0, mark=square*, mark size=2pt]
table [row sep=\\]{%
25 0.372\\
50 0.493\\
100 0.548\\
150 0.563\\
200 0.566\\
};
\end{axis}
\end{tikzpicture}
\caption{BoxE validation performance over YAGO3-10 versus dimensionality.} %
\label{fig:robustness}
\end{wrapfigure}

\subsection{Robustness experiment}
\label{app:robustness}
In this experiment, we evaluate the dependence of BoxE on dimensionality $d$, to understand its prospective performance in a computationally restricted setting. 

\paragraph{Experimental setup.} We train BoxE with uniform negative sampling on YAGO3-10 using $d=\{25,50,100,150,200\}$. We only tune the margin and fix the learning rate, batch size, and number of negative samples to $10^{-3}$, 4096, and 150, respectively. $L2$ norm and data augmentation are used across all experiments. We report peak MRR recorded over the validation set. The final margins were $\gamma=6$ for $d=25$ and $\gamma=10.5$ otherwise.

\paragraph{Results.} A plot of validation MRR versus dimensionality is drawn in \Cref{fig:robustness}. BoxE maintains very strong performance, even at $d=50$, rivaling that of state-of-the-art translational model RotatE, even with just uniform negative sampling. Furthermore, it performs at near-optimal level with $d=100$, and is already state-of-the-art on YAGO3-10 at this small dimensionality. Hence, BoxE proves to be very robust for performing knowledge base completion with restricted computational power. 

\subsection{Box volume information for BoxE following training on YAGO3-10}
\label{app:YAGOVolumes}
\begin{wraptable}[43]{r}{0.45\textwidth}
	\centering
	\caption{Geometric mean volume per dimension for all relation boxes in YAGO3-10 following training.} 
	\label{tab:YAGOVols} 
	\begin{tabular}{l@{\hskip 6pt}c@{\hskip 3pt}c@{\hskip 3pt}}
		\toprule 
		Relation & Head Box & Tail Box\\
		\cmidrule{2-3}
		$\mathsf{actedIn}$ & 0.456 & 0.479\\
		$\mathsf{created}$ & 0.966 & 0.905\\
		$\mathsf{dealsWith}$ & 0.373 & 0.366\\
		$\mathsf{diedIn}$ & 0.383 & 0.480\\
		$\mathsf{directed}$ & 0.474 & 0.461\\
		$\mathsf{edited}$ & 0.461 & 0.441\\
		$\mathsf{exports}$ & 0.238 & 0.260 \\
		$\mathsf{graduatedFrom}$ & 0.608 & 0.526 \\
		$\mathsf{happenedIn}$ & 0.453 & 0.363\\
		$\mathsf{hasAcademicAdvisor}$ & 0.655 & 0.605 \\
		$\mathsf{hasCapital}$ & 0.390 & 0.347\\
		$\mathsf{hasChild}$ & 0.299 & 0.761\\
		$\mathsf{hasCurrency}$ & 0.228 & 0.239\\
		$\mathsf{hasGender}$ & 0.669 & 0.688 \\
		$\mathsf{hasMusicalRole}$ & 0.328 & 0.427 \\
		$\mathsf{hasNeighbor}$ & 0.311 & 0.312\\
		$\mathsf{hasOfficialLanguage}$ & 0.213 & 0.255\\
		$\mathsf{hasWebsite}$ & 0.159 & 0.143 \\
		$\mathsf{hasWonPrize}$ & 0.264 & 0.381\\
		$\mathsf{imports}$ & 0.249 & 0.241 \\
		$\mathsf{influences}$ & 0.510 & 0.567\\
		$\mathsf{isAffiliatedTo}$ & 0.257 & 0.557\\
		$\mathsf{isCitizenOf}$ & 0.544 & 0.614 \\
		$\mathsf{isConnectedTo}$ & 0.403 & 0.388\\
		$\mathsf{isInterestedIn}$ & 0.644 & 0.496\\
		$\mathsf{isKnownFor}$ & 0.632 & 0.623\\
		$\mathsf{isLeaderOf}$ & 0.446 & 1.005\\
		$\mathsf{isLocatedIn}$ & 0.496 & 0.547  \\
		$\mathsf{isMarriedTo}$ & 0.923 & 0.924\\
		$\mathsf{isPoliticianOf}$ & 0.361 & 0.521\\
		$\mathsf{livesIn}$ & 0.536 & 0.341 \\
		$\mathsf{owns}$ & 0.907 & 0.485\\
		$\mathsf{participatedIn}$ & 0.389 & 0.471 \\
		$\mathsf{playsFor}$ & 0.284 & 0.469\\
		$\mathsf{wasBornIn}$ & 0.465 & 0.445\\
		$\mathsf{worksAt}$ & 0.498 & 0.488\\
		$\mathsf{wroteMusicFor}$ & 0.450 & 0.646\\
		\bottomrule
	\end{tabular}
\end{wraptable}

In \Cref{tab:YAGOVols}, we report the geometric mean of box volume across dimensions for all head and tail boxes for the 37 relations in YAGO3-10 following training. 
These numbers are computed from the same configuration whose results are reported in the main paper for BoxE(u). Note that, as explained in \Cref{app:exp}, boxes are mapped to the space $[-1,1]^d$ using the hyperbolic tangent function, so the geometric mean volume is upper-bounded by 2. From \Cref{tab:YAGOVols}, we can make the following four very interesting observations: 

First, we see that more popular relations, in terms of entities they connect, tend to be represented with larger boxes in the embedding space. This confirms our intuition that the boxes effectively define entity classes, and thus larger classes, are met with larger boxes in the embedding space. 
For example, the less popular relation $\mathsf{hasWebsite}$ has very small boxes of mean volume about 0.15, as it is only makes up 68 facts in the YAGO training dataset. By contrast, the relation $\mathsf{created}$ has both boxes with mean volume above 0.9, and appears in over 1,400 facts. 

Second, we observe that the size of relation boxes also correlates with implicit entity types, in addition to relation popularity.
Indeed, the relation $\mathsf{playsFor}$, despite appearing over 300,000 times, only has box volumes 0.284 and 0.469 respectively, whereas $\mathsf{isLeaderOf}$, with less than 1,000 facts, has a tail box of mean volume exceeding 1. This is due to the diversity in entity types appearing at these relations: For $\mathsf{playsFor}$, head entities are athletes, which cluster together in a smaller region of the embedding space, and tail entities are football/sports clubs, which are more diverse, but still quite similar semantically. By contrast, head entities for $\mathsf{isLeaderOf}$ are individuals, with medium variability, but tail entities can be anything from very different countries (e.g., Mali, Kuwait) to cities, districts, and towns (e.g., Toronto, Oxnard (California)), to political parties and associations (e.g., Democratic Governors Association, Hungarian Communist Party), which are vastly different types of entities, and this results in an extremely large tail box for $\mathsf{isLeaderOf}$. 

Third, we observe that relative box sizes accurately reflect the type of their underlying relation. More specifically, larger tail boxes tend to denote one-to-many relations, larger head boxes indicate a many-to-one relation, and similar sizes indicate many-to-many or one-to-one relations. This is especially evident for the one-to-many relations $\mathsf{hasChild}$ (0.299 vs 0.761), and $\mathsf{isAffiliatedTo}$ (0.257 vs 0.557), and for many-to-one relations $\mathsf{isInterestedIn}$ (0.644 vs 0.496), and $\mathsf{graduatedFrom}$ (0.608 vs 0.526). 

Finally, we note that symmetric relations in YAGO3-10, namely $\mathsf{hasNeighbor}$ and $\mathsf{isMarriedTo}$, are represented with near-identically sized boxes. This is a very important finding, as it indicates that BoxE succesfully captures the symmetry inference pattern, for which a necessary condition is having identical head and tail boxes. 

All in all, these results further highlight the interpretability of BoxE, in terms of capturing inference patterns, accurately inferring and portraying entity classes, and inferring and successfully modelling relation types, which other models are unable to achieve. 

\subsection{Rule injection experiment}
\label{app:ruleInjExp}
\begin{wrapfigure}[17]{r}{0.5\textwidth}
\centering
\begin{tikzpicture}[scale=0.75]
\definecolor{color0}{rgb}{0.83921568627451,0.152941176470588,0.156862745098039}
\definecolor{color1}{rgb}{0.172549019607843,0.627450980392157,0.172549019607843}
\pgfplotsset{every tick label/.append style={font=\Large}}
\begin{axis}[
legend cell align={left},
legend entries={{\Large BoxE},{\Large BoxE + RI}},
legend style={at={(0.51,0.23)}, font=\Large, line width=0.4mm, anchor=north west, draw=white!80.0!black, label style={font=\large}},
tick align=outside,
tick pos=left,
x grid style={white!69.01960784313725!black},
xlabel={\Large Training Epoch},
 x label style={at={(axis description cs:0.5,-0.05)},anchor=north},
y label style={at={(axis description cs:0,.5)},anchor=south},
xmajorgrids,
xmin=00, xmax=2000,
ylabel={\Large Validation MRR},
ymajorgrids,
ymin=0, ymax=1
]
\addlegendimage{no markers, color0}
\addlegendimage{no markers, color1}
\addplot [line width=0.6mm, color0]
table [row sep=\\]{%
0   0 \\
50 0.272 \\
100 0.287 \\
150 0.299 \\
200	0.315 \\
400	0.370 \\
600	0.439 \\
800	0.484 \\
1000 0.521 \\
1200 0.538 \\
1400 0.554\\
1600 0.562 \\
1800 0.568 \\
2000 0.574  \\
};
\addplot [line width=0.6mm, color1]
table [row sep=\\]{%
0 0 \\
50 0.785 \\
100 0.891 \\
150 0.924 \\
200	0.952 \\
400	0.973 \\
600	0.978 \\
800	0.979 \\
1000 0.979 \\
1200 0.978 \\
1400 0.978 \\
1600 0.977 \\
1800 0.977 \\
2000 0.976 \\
};
\end{axis}
\end{tikzpicture}
\caption{BoxE and BoxE+RI learning curves.} %
\label{fig:learningCurve}
\end{wrapfigure}
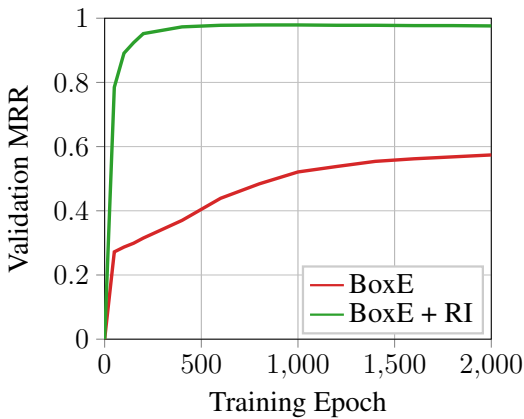
In this section, we provide additional information about the rule injection experiment presented in the paper. In particular, we give a more complete presentation of model convergence with and without rule injection, and provide further details on SportsNELL. 

\paragraph{Learning curves of BoxE, BoxE+RI.} The learning curves of BoxE, and Box+RI, defined with MRR as the performance metric, across the 2000 training epochs of the rule injection experiment, is shown in \Cref{fig:learningCurve}. The two curves highlight a remarkable improvement stemming from injecting the SportsNELL ontology. Indeed, BoxE+RI converges to peak performance within 500 epochs, and mostly stablises its peak MRR following this point, whereas standard BoxE does not fully converge, even after the whole 2000 epochs have elapsed. Furthermore, the difference in performance between these two models is very significant. Hence, rule injection not only yields better-performing KBC systems, but also enables faster, more reliable training of these systems.  

\paragraph{Further details about SportsNELL.}
\label{app:SportsNELL}
SportsNELL initially consists of 181,936 facts, 11 relations and 4,252 sports-related entities, such that all its entities initially appear 50 or more times in NELL across these 11 relations. Its \emph{logical closure} w.r.t the SportsNELL ontology is then computed., i.e., ontology rules are repeatedly applied to deduce new facts until no new facts can be deduced: new facts in the deductive closure are direct results of rule application, and thus their correct prediction indicates a good capturing of the underlying ontology. The resulting combined dataset, referred to as $\text{SportsNELL}^\text{C}$, contains a total of 326,650 facts.

\end{document}